\pdfoutput=1
\documentclass[11pt]{article}

\usepackage[final]{acl}

\usepackage{times}
\usepackage{latexsym}
\usepackage{graphicx}
\usepackage{multirow}
\usepackage{booktabs}
\usepackage{adjustbox}
\usepackage{algorithm}
\usepackage{algpseudocode}
\usepackage{tabularx}
\usepackage{microtype}
\usepackage{amsmath}
\usepackage{amssymb}
\newcommand{\jr}[1]{}
\newcommand{\mh}[1]{}

\newcommand{\Description}[2][]{}

\usepackage[T1]{fontenc}

\usepackage[utf8]{inputenc}

\usepackage{microtype}

\usepackage{inconsolata}

\usepackage{graphicx}

\title{TEMPS: Temporal Sentence Embeddings for Temporal Information Retrieval}

\author{
  \textbf{Mourad Hassani\textsuperscript{1,2}},
  \textbf{Julien Romero\textsuperscript{1}},
  \textbf{Amel Bouzeghoub\textsuperscript{1}},
  \textbf{Christian Jacquelinet\textsuperscript{2}}
\\
\\
  \textsuperscript{1}SAMOVAR, Télécom SudParis, Institut Polytechnique de Paris, France,
  \textsuperscript{2}Aldebaran Care, France
\\
  \small{
    \textbf{Correspondence:} \href{mailto:mourad.hassani@telecom-sudparis.eu}{mourad@aldebaran.care}
  }
}

\begin{document}
\maketitle
\begin{abstract}
Modern information retrieval (IR) systems rarely represent time, yet many information needs depend on it: in clinical, journalistic, and legal search, \emph{when} an event occurred can decide whether a document is relevant. Dense retrievers and Retrieval-Augmented Generation (RAG) pipelines match queries to documents well on topic but poorly on time, so they surface content that is on-topic yet temporally wrong. We introduce \textbf{Temporal Textual Similarity (TTS)}, a task that measures how well two anchored texts align in time, independent of their topical similarity. We then present \textbf{TEMPS} (Temporal Embedding Model for Precise Search), a modular temporal branch that attaches to a frozen semantic retriever and trains on that signal. It resolves anchored temporal expressions to intervals and moment-matches each one to a Gaussian; the resulting ordering supervises an anchor-date-conditioned encoder, whose score we fuse with the semantic score at inference. Grounding supplies the supervision, so training uses no hand-labeled temporal data. The temporal score itself is the Gaussian-KL inclusion measure from distributional embeddings; what TEMPS adds is the grounding and the moment-matched supervision. On three temporal benchmarks, TEMPS improves MRR for every semantic backbone tested and, on TS-Retriever, lifts R@1 from 19.92 to 25.39 over the prior temporal state of the art.
\end{abstract}

\section{Introduction}
\label{sec:introduction}
Information Retrieval (IR) underpins many Natural Language Processing (NLP) applications, including information extraction, topic detection, and summarization. Traditional IR has largely focused on \textit{semantic} relevance, matching documents to queries by topical similarity~\cite{campos2014survey, rizzo2022ranking}, yet many information needs are also inherently \textit{temporal}~\cite{alonso2011temporal}: \emph{when} something happened can matter as much as \emph{what} happened. Many documents carry timestamps, and their text often anchors events in time as
well, whether with an absolute expression (e.g., ``on January 15th, 2008'') or a
deictic one resolved against that timestamp (e.g., ``last summer'')~\cite{derczynski2012massively};
interpreting these contexts is essential for results that are both topically and
temporally relevant~\cite{campos2014survey, han-etal-2025-temporal}.

\begin{figure}[h]
    \centering
    \includegraphics[width=\linewidth]{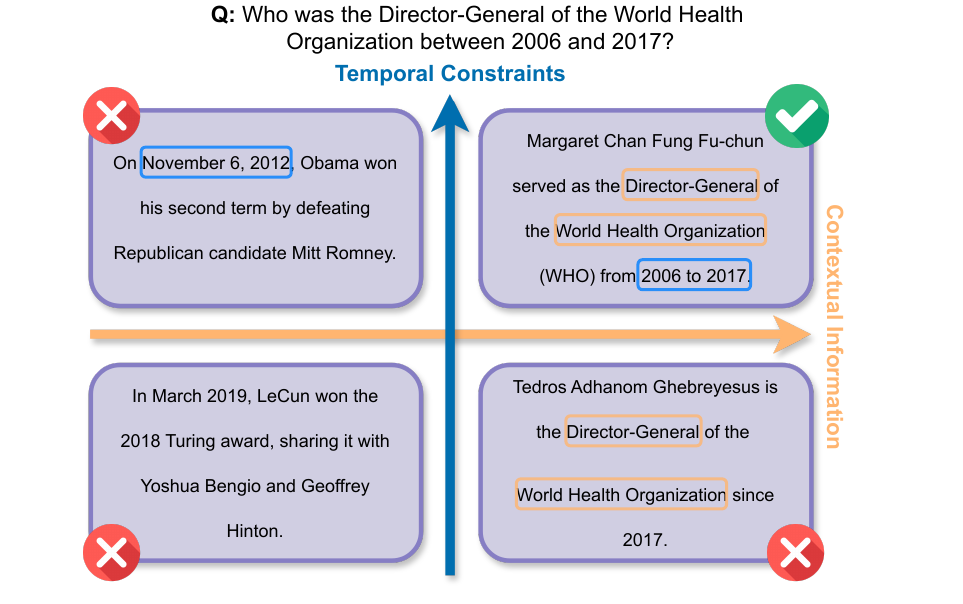}
    \Description[Topical relevance and temporal compatibility are complementary]{Topical relevance and temporal compatibility are complementary}
    \caption{Topical relevance and temporal compatibility are complementary. Only
    the top-right passage satisfies both; the bottom-right passage is topically
    relevant but incompatible with the requested period, showing why semantic
    relevance alone can fall short.}
    \label{fig:intro-example}
\end{figure}

Failing to account for temporal constraints can lead to factually correct but contextually misleading answers. Figure~\ref{fig:intro-example} illustrates this: given the query ``Who was the Director-General of the World Health Organization between 2006 and 2017?'', two passages appear semantically relevant, yet only one aligns with the correct time frame. The passage directly below the correct one is just as topical but covers a tenure outside the requested window, and a retriever scoring only topical relevance cannot separate them. Without temporal reasoning, IR systems risk returning outdated, premature, or otherwise temporally mismatched information, undermining trust in downstream applications such as search, question answering, and decision support.

This need spans many domains: clinical narratives, where diagnoses and treatments unfold over time~\cite{moharasar2016semi}; journalism, where ignoring a publication date recirculates outdated stories as misinformation~\cite{musi2022staying}; and legal contexts, where a document's validity depends on its date of enactment. Even consumer search for reviews or events relies on temporal alignment to stay relevant.

Historically, retrieval models such as Term Frequency–Inverse Document Frequency (TF-IDF)~\cite{salton1988term} and BM25~\cite{10.1561/1500000019} operated solely on lexical overlap, with no mechanism for temporal reasoning~\cite{qian-etal-2024-timer4}. Dense retrieval models~\cite{karpukhin2020dense, khattab2020colbert, gao2021simcse}
improved semantic matching by encoding queries and documents into a shared vector
space with pretrained transformer encoders. They are now the standard first stage
of Retrieval-Augmented Generation (RAG)~\cite{lewis2020retrieval}, where the
retrieved passages are passed to a generator that synthesizes an answer.

However, dense retrievers inherit their representations from encoders such as
BERT~\cite{devlin2018bert}, ModernBERT~\cite{warner2024smarter}, and
ELECTRA~\cite{clark2020electra}, which are not trained to make temporal
distinctions. Time-sensitive benchmarks make the cost visible: state-of-the-art
long-document QA systems reach 46\% accuracy on TimeQA against 87\% for
humans~\cite{chen2021dataset}, and generative LLMs show the same
weakness~\cite{song2025bridging}. This stems from two factors. First, the corpora used to pretrain these encoders underrepresent explicit
temporal expressions and rely on coarse-grained timestamps, leaving models
ill-equipped for questions involving sequence, overlap, and
duration~\cite{wang2023bitimebert}. Second, the embedding spaces of dense retrievers are static snapshots: they encode knowledge available at training time but cannot adapt to evolving timelines~\cite{sojitra2024timeline} or reason about time-dependent constraints. The result is \emph{temporally brittle} retrieval that matches on semantics while missing time-sensitive relationships, returning temporally incorrect documents.

Temporal cues compound this problem, since they are often implicit or ambiguous. Resolving ``earlier this year'' requires the document's publication date, while ``five years from now'' requires arithmetic over an anchor; without dedicated modeling, such cases yield irrelevant results.

A central difficulty is that temporal compatibility is often \emph{asymmetric}. A passage about ``in 2020'' is compatible with a query about a specific date that year, but a passage about a single day should not be equally compatible with a query about the whole year. Symmetric functions such as cosine similarity or interval overlap cannot express this directionality, scoring ``day contained in year'' and ``year contains day'' identically. This motivates modeling temporal expressions as distributions, where the mean captures location, the variance captures granularity or uncertainty, and a directional divergence distinguishes fine-to-coarse compatibility from its reverse.

We address these limitations with a framework that models temporal information
explicitly and folds it into dense retrieval without retraining the underlying
retriever. The core idea is to ground anchored temporal language to intervals
and derive supervision directly from those intervals, then train an
anchor-date-conditioned module to reproduce it and fuse with any semantic
retriever. To represent and compare intervals, we adopt the asymmetric Gaussian-KL scoring
used in prior distributional embedding work~\cite{vilnis2014word,
he2015learning, yoda2023sentence}: its directional divergence gives us the
fine-to-coarse asymmetry that temporal inclusion needs; what is new is the
temporal grounding and the moment-matched interval-to-Gaussian supervision, not
the divergence.

Our contributions are fourfold:
\begin{itemize}
\itemsep0em
    \item We formulate \textbf{Temporal Textual Similarity (TTS)}, a task that
    scores whether two anchored texts are temporally compatible, independent of
    topical similarity, giving STS (Semantic Textual Similarity) a temporal counterpart that existing
    similarity resources do not capture.
    \item We show this compatibility can be \textbf{trained at scale without
    manual labeling}: a pipeline grounds temporal language to intervals and
    moment-matches them into directional Gaussian targets, yielding 60M training
    tuples.
    \item We propose a new framework called \textbf{TEMPS} (Temporal Embedding Model for Precise Search) that combines
    \emph{frozen} semantic retrievers with fine-grained temporal reasoning
    at a minimal cost. It repurposes
    established Gaussian-KL inclusion machinery for temporally grounded text
    rather than introducing a new divergence.
    \item We isolate the learned encoder with a rule-based control that swaps
    temporal grounding into TEMPS, and quantify where it helps and where
    it does not.
\end{itemize}

We provide the code, trained TEMPS module, training-data generation pipeline, and all evaluation scripts at
\url{https://github.com/aldebaran-care/TEMPS}.

\section{Preliminaries}

\paragraph{Temporal Information \& Expressions} 
Temporal information facilitates time-aware retrieval through computable relations~\cite{allen1983maintaining}, hierarchical granularity, and normalizable expressions~\cite{pustejovsky2003timeml}. These expressions are categorized by their grounding~\cite{derczynski2012massively}: \textit{absolute} (explicit timestamps like ``January 15th''), \textit{deictic} (speaker-relative, e.g., ``yesterday''), \textit{anaphoric} (relative to a prior reference), and \textit{duration} (defined spans).

\paragraph{TimeML Specification} 
TimeML~\cite{pustejovsky2003timeml, pustejovsky2005temporal, pustejovsky2010iso} is the standard markup for annotating these expressions in natural language~\cite{jia2018tempquestions}, forming the foundation for many temporal taggers~\cite{llorens2010tipsem, strotgen2010heideltime, lee2014context}. Instead of a standalone table, we outline our synthetic dataset's TimeML annotations here: Dates (e.g., \texttt{2024-09} for September 2024), Offsets (\texttt{THIS P1D} for the present day), References (\texttt{PRESENT\_REF} for now), and Intervals (\texttt{2023-SU, 2024-FA} for summer 2023 to fall 2024).

\paragraph{Scope of this work}
TEMPS covers expressions that can be \emph{anchored and resolved} to an interval: absolute dates, deictic expressions and offsets resolved against the anchor date, \texttt{PRESENT\_REF}-style references, seasons, durations, and intervals. It does not perform discourse-level reasoning: references needing an unobserved antecedent, unanchored vague ordering, and event-event relations lie outside what the supervision teaches, and the pipeline inherits the tagger's misses.

\section{Related Work}

\paragraph{Temporal Information Retrieval.}
Temporal Information Retrieval (TIR) incorporates time into search by exploiting document timestamps, temporal expressions, and explicit or implicit temporal intent in queries~\cite{campos2014survey, kanhabua2016temporal}. This allows systems to rank documents by both topical and temporal relevance, which is essential for dynamic collections where the same entity or event may have different correct interpretations at different times.

\paragraph{Semantic Textual Similarity.}
Semantic Textual Similarity (STS) measures the degree of semantic equivalence
between sentence pairs and supports tasks such as machine translation,
summarization, QA, and semantic search~\cite{agirre-etal-2012-semeval,
agirre-etal-2016-semeval, cer-etal-2017-semeval}. The STS
Benchmark~\cite{cer-etal-2017-semeval} and SICK~\cite{marelli-etal-2014-sick}
provide human similarity and relatedness annotations respectively, but both
target semantic relatedness rather than temporal alignment. As a result, high
STS scores do not necessarily imply that two sentences express compatible
temporal information.

\paragraph{Time-Sensitive Question Answering.}
Time-sensitive QA requires models to answer under temporal constraints, e.g.,
distinguishing facts that hold before or after a given date. Benchmarks such as
TimeQA~\cite{chen2021dataset} test whether models can locate and reason over
explicit and implicit temporal expressions in long documents in order to produce
answers valid for the specified time context, where the best reported systems
reach 46\% accuracy against 87\% for humans.

\paragraph{Retrieval-Augmented Generation.}
RAG improves factuality in knowledge-intensive tasks by retrieving external
evidence before generation~\cite{lewis2020retrieval}, which matters most for
facts that are rare in pretraining data~\cite{kandpal2023large}. Modern systems commonly rely on sparse retrievers such as BM25~\cite{10.1561/1500000019} and on
embedding models such as E5~\cite{e5} and BGE-M3~\cite{chen2024bge}, the latter
combining dense, sparse, and multi-vector scoring.

\paragraph{Gaussian Embeddings and Directional Similarity.}
Gaussian embeddings represent items as distributions rather than points, allowing the mean to encode location and the covariance to encode uncertainty or generality. Vilnis and McCallum~\cite{vilnis2014word} use KL divergence between Gaussian word embeddings to model asymmetric inclusion in lexical semantics; KG2E~\cite{he2015learning} applies Gaussian embeddings and asymmetric KL to directional knowledge-graph relations; and GaussCSE~\cite{yoda2023sentence} uses a bounded inverse-KL score for entailment-oriented sentence similarity. TEMPS follows this lineage rather than introducing a new Gaussian divergence or similarity head. Its novelty is temporal: grounded intervals are converted into Gaussian supervision by moment matching, and an anchor-date-conditioned branch learns to reproduce this temporal compatibility signal for retrieval.

\paragraph{Temporal Retrieval for RAG.}
TS-Retriever and TSContriever~\cite{wu2024time} directly address time-sensitive
retrieval: the benchmark pairs semantically similar but temporally distinct
documents, and the model fine-tunes Contriever~\cite{contriever} on contrastive
pairs whose negatives are built by perturbing the time specifier in the question
while holding the passage fixed. This line of work shows the value of temporal supervision, but existing approaches remain tied to specific retriever training setups. In contrast, our work introduces TTS as a temporal similarity objective and TEMPS as a modular temporal component that can be combined with different semantic retrievers.

\paragraph{Gap.}
Despite progress in STS, TIR, and RAG, current embedding models remain limited on temporally rich text. Standard similarity functions such as cosine similarity are symmetric and cannot naturally represent asymmetric relations such as a specific date being contained in a broader interval. Moreover, existing STS resources do not teach models to ignore topical similarity when temporal information conflicts. These gaps motivate temporally aware representations and datasets that explicitly model temporal alignment.

\section{Problem Definition}
\label{sec:problem}

We propose the \textbf{Temporal Textual Similarity (TTS)} task to improve time-aware \textbf{IR} models. Designed as a temporal analogue of the STS task, TTS trains embedding models to measure the alignment of temporal information expressed in sentences.

However, unlike STS, which assesses semantic or topical relatedness, \textbf{TTS disregards semantic similarity and focuses solely on the temporal dimension}.

Formally:

\begin{quote}
\textit{Given a quadruple \((s_1, d_1, s_2, d_2)\), where \(d_1\) and \(d_2\) are anchor dates for interpreting temporal expressions in the texts \(s_1\) and \(s_2\), estimate the degree to which the texts express similar temporal information, independent of semantic or topical similarity.}
\end{quote}

This task is intended as a basis for developing retrieval systems sensitive to temporal context.

\paragraph{Unit of text.} Training pairs are short anchored expressions, but retrieval operates on passages, so $s_i$ ranges over passages of up to 512 tokens in our work rather than single sentences; Section~\ref{sec:system-architecture} describes how a passage with several temporal expressions, or none, is reduced to one Gaussian.

\section{Methodology}
\label{sec:methodology}
TEMPS learns a temporal signal that complements semantic similarity. The method
has two parts: (i) weakly supervised temporal labels for pairs of anchored
expressions, and (ii) a lightweight temporal embedding module that can be
combined with arbitrary dense retrievers. Both the labels and the inference-time
score use the same Gaussian-KL form, so training reduces to teaching the module
to reproduce a closed-form temporal target. Detailed dataset construction and
model/training specifications are provided in
Appendices~\ref{sec:appendix-dataset} and~\ref{sec:appendix-model-details}.

\subsection{Temporal Supervision}

\paragraph{Two Gaussians, two jobs.} ``Gaussian'' does double duty below. On the
\emph{label side}, a grounded interval $I=[a,b]$ is read as a uniform distribution over its
inclusive day range and replaced by the Gaussian with the same mean and variance
(Eq.~\ref{eq:moment-match}): its axis is the calendar and its variance is fixed
by the interval width. We are not claiming event times are normally distributed;
the Gaussian is a tractable stand-in, chosen because the KL between two interval
uniforms is degenerate (Appendix~\ref{sec:moment-matching}). On the \emph{model
side}, TEMPS emits a diagonal Gaussian in a learned latent space whose
coordinates are not calendar days and whose variance is learned. The CoSENT~\cite{10.1109/TASLP.2024.3402087}
objective ties the two together by training the model-side similarities to
reproduce the ordering the label-side Gaussians impose; the spaces need not
coincide, only their rankings.

Each training instance is a tuple $(s_1,d_1,s_2,d_2,ts)$, where $s_i$ is a text
containing a temporal expression, $d_i$ is its anchor date at day granularity,
and $ts$ is a temporal similarity value. We generate these tuples from synthetic
TimeML-style expressions spanning the years 1000-2030 and the English portion of
the Multilingual MLM Temporal Tagging Resources dataset~\cite{lange2022multilingual},
yielding roughly 60M tuples in total.

Temporal labels are computed by grounding expressions to intervals and applying a
directional temporal similarity function. Given intervals $I_1=[a_1,b_1]$ and
$I_2=[a_2,b_2]$ for the temporal expressions in $s_1$ and $s_2$, respectively, we
first represent each interval as a Gaussian by moment matching the uniform
distribution over its inclusive day range:
\begin{equation}
\begin{aligned}
p_I &= \mathcal{N}(\mu_I,\sigma_I^2), \\
\mu_I &= \frac{a+b+1}{2}, \\
\sigma_I^2 &= \frac{(b-a+1)^2}{12}.
\end{aligned}
\label{eq:moment-match}
\end{equation}
The mean captures temporal location, and the variance captures the granularity or
uncertainty of the expression: a single day has low variance, whereas a month or
year has higher variance. The distributional view matters because temporal
relevance is not symmetric. A specific date can be judged compatible with a
broader containing interval, while the reverse direction should be penalized.

We therefore define the temporal similarity function as an asymmetric KL
divergence from the first interval to the second:
\begin{equation}
\begin{aligned}
\mathrm{TSF}(I_1 \!\to\! I_2)
&= D_{\mathrm{KL}}(p_{I_1} \Vert p_{I_2}) \\
&= \frac{1}{2}\log\frac{\sigma_2^2}{\sigma_1^2} \\
&\quad + \frac{\sigma_1^2 + (\mu_1 - \mu_2)^2}{2\sigma_2^2}
 - \frac{1}{2}.
\end{aligned}
\end{equation}
Smaller values indicate stronger temporal compatibility in the $s_1\!\to\!s_2$
direction. For training we use the bounded value
$ts=1/(1+\mathrm{TSF}(I_1\!\to\!I_2))$, matching the inverse-KL form used in prior
Gaussian sentence embeddings~\cite{yoda2023sentence}. This gives smooth,
directional supervision and keeps the training labels in the same form as the
Gaussian embedding head used at inference. Additional derivation and examples are
given in Appendix~\ref{sec:tsf}.

\subsection{Temporal Model}
\label{sec:system-architecture}

\begin{figure}[ht]
    \centering
    \includegraphics[width=\linewidth]{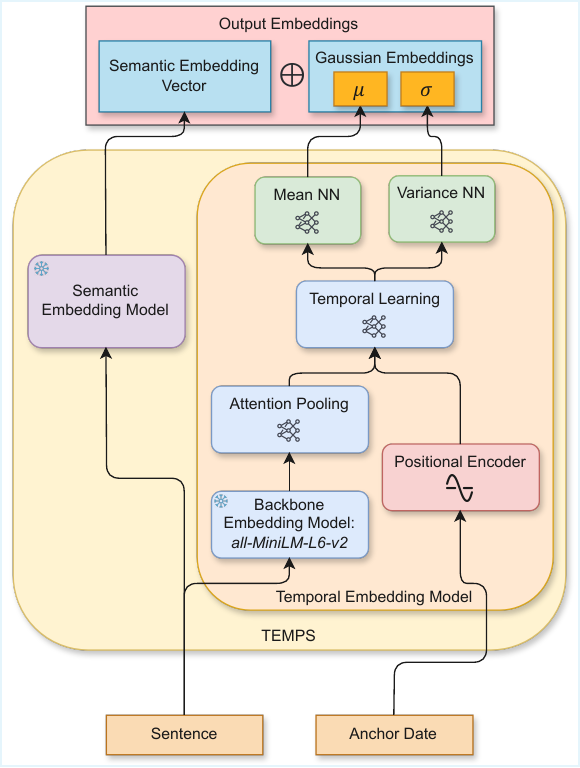}
    \Description[TEMPS' Model Architecture]{TEMPS' Model Architecture}
    \caption{TEMPS' Model Architecture}
    \label{fig:system-architecture}
\end{figure}

\paragraph{Architecture.}
Figure~\ref{fig:system-architecture} shows TEMPS, which consists of a semantic branch and a temporal branch. The semantic branch supplies the standard dense representation used for topical relevance. The temporal branch receives a passage and an anchor date, encodes the passage with a frozen all-MiniLM-L6-v2 backbone at up to 512 tokens, encodes the date with a sinusoidal positional representation, and maps the combined representation to a Gaussian embedding. Following prior Gaussian embedding models~\cite{vilnis2014word, yoda2023sentence}, the mean captures temporal location, while the diagonal variance captures temporal granularity and uncertainty.

The passage, not the sentence, is the unit at inference. A passage mentioning several dates is not split: attention pooling weights the temporal tokens and emits one Gaussian for the whole passage, so a wide range of dates yields a correspondingly wide variance. Passages with no temporal expression carry no interval, so they never appear in temporal training and their Gaussians are uncalibrated; the semantic branch carries these cases, and $\alpha$ bounds how far an uncalibrated temporal score can move them.

\paragraph{Temporal Similarity.}
For an anchored text $(s,d)$, TEMPS outputs a semantic vector $E_S$ and temporal Gaussian parameters $(\mu,\sigma^2)$. Given two inputs, we combine semantic cosine similarity with an asymmetric temporal score $s_{1,2}$:

\begin{equation}
\begin{aligned}
s_{1,2}
&= (1-\alpha)\,\cos(E_{S,1}, E_{S,2}) + \alpha\,\tau_{1,2}, \\
\tau_{1,2}
&= \frac{1}{1 + D_{\mathrm{KL}}(p_1 \Vert p_2)}, \\
p_j
&= \mathcal{N}(\mu_j,\mathrm{diag}(\sigma_j^2)),\quad j \in \{1,2\}.
\end{aligned}
\end{equation}

Here, $\alpha \in [0,1]$ controls the temporal contribution; the two endpoints name the ablations we report, with $\alpha=0$ the semantic-only baseline and $\alpha=1$ the temporal-only variant. Scores from the two branches are min-max normalized per query before fusion. Because $D_{\mathrm{KL}}$ is directional, the score can distinguish relations such as a date being contained in a broader interval from the reverse direction.

\paragraph{Training and Integration.}
The temporal branch is trained on TTS supervision with a CoSENT-style ranking loss, while the semantic branch can be replaced by any external embedding model. This modular design lets TEMPS add temporal awareness to existing retrievers without retraining their semantic encoders. Architectural, loss, and hyperparameter details are provided in Appendix~\ref{sec:appendix-model-details}.

\section{Experiments}
\label{sec:experiments}

\subsection{Experimental Setup}

We evaluate TEMPS on three temporal retrieval benchmarks: \textbf{TimeQA}~\cite{chen2021dataset}, which tests evolving facts over long documents; \textbf{TempReason}~\cite{tan2023towards}, which evaluates date arithmetic and event-time grounding; and \textbf{TS-Retriever}~\cite{wu2024time}, which stresses retrieval among semantically similar but temporally mismatched documents. We compare against lexical, rule-based, dense, LLM-based, and temporal retrieval baselines, including BM25, SUTime~\cite{chang2012sutime}, MPNet\textsubscript{base}-v2~\cite{mpnet}, E5\textsubscript{base}-v2~\cite{e5}, BGE\textsubscript{large}-v1.5~\cite{bge}, Mistral, and TSContriever~\cite{wu2024time}. For each dense retriever, we also report the corresponding +TEMPS variant and a rule-based +SUTime control described below. To test whether temporal retrieval gains transfer to an end task, we additionally evaluate a TimeQA RAG pipeline that retrieves the top-5 passages and uses Qwen2.5-7B-Instruct~\cite{qwen2024qwen25} as the generator.

For retrieval, we report MRR, NDCG@5, Recall@5, and Precision@5. For downstream RAG, we report only answer-level metrics: exact match (EM), token-level F1, and answer containment. The temporal-semantic interpolation parameter $\alpha$ is a benchmark-level calibration parameter, not a trained model parameter, and we select it on a held-out split. Each benchmark is partitioned at random into 20\% validation and 80\% test; one $\alpha$ per benchmark is chosen from validation MRR alone and then held fixed across every backbone, metric, and significance test, and the test split is scored once. This selects $\alpha=0.6$ for TimeQA, $0.8$ for TempReason, and $0.3$ for TS-Retriever. Every number reported below is a test-split number. Additional benchmark statistics, baseline descriptions, and computational details are provided in Appendix~\ref{sec:appendix-experimental-details}.

\paragraph{Rule-based control.} The \textbf{+ SUTime} rows are an ablation that changes one component of the +TEMPS pipeline, the source of the temporal Gaussian: SUTime grounds the expressions and Eq.~\ref{eq:moment-match} maps the intervals, with fusion, normalization and $\alpha$ held fixed (Appendix~\ref{sec:appendix-experimental-details}). Any difference between a \textit{+ TEMPS} row and its \textit{+ SUTime} row measures the temporal representation alone.

\subsection{Results}

\begin{table*}[!t]
\centering
\caption{Retrieval results on the held-out test split, with $\alpha$ selected per benchmark on a 20\% validation split from MRR alone. Unaugmented backbones are the $\alpha=0$ endpoint and \textit{TEMPS w/o Semantic Model} the $\alpha=1$ endpoint; \textit{+ SUTime} is the rule-based control (same fusion, normalization and $\alpha$, temporal Gaussian from SUTime grounding). Gains are relative to the unaugmented backbone; best per benchmark and metric in \textbf{bold}.}
\label{tab:model_performance}

\footnotesize
\setlength{\tabcolsep}{3pt}
\renewcommand{\arraystretch}{0.90}

\begin{adjustbox}{
    max width=\textwidth,
    max totalheight=0.88\textheight,
    keepaspectratio
}
\begin{tabular}{llcccccc}
\toprule
\textbf{Benchmark} & \textbf{Model} & \textbf{MRR} & \textbf{NDCG@5} & \textbf{Recall@5} & \textbf{Precision@5} & \textbf{MRR Gain} & \textbf{\# params} \\
\midrule

\multirow{15}{*}{\texttt{TimeQA ($\alpha=0.6$)}}
& BM25 & 0.2398 & 0.2472 & 0.4059 & 0.0812 & - & - \\
& SUTime & 0.4704 & 0.4966 & 0.6471 & 0.1294 & - & - \\
& TEMPS w/o Semantic Model & 0.4587 & 0.4830 & 0.6366 & 0.1273 & - & 27M \\
& $MPNet_{base}$-v2 & 0.3008 & 0.3121 & 0.4803 & 0.0961 & - & 109M \\
& $MPNet_{base}$-v2 + TEMPS & 0.5043 & 0.5320 & 0.6915 & 0.1383 & +67.65\% & 136M \\
& $MPNet_{base}$-v2 + SUTime & 0.2758 & 0.2905 & 0.4718 & 0.0944 & -8.32\% & 109M \\
& $E5_{base}$-v2 & 0.3334 & 0.3712 & 0.6028 & 0.1206 & - & 109M \\
& $E5_{base}$-v2 + TEMPS & 0.5339 & \textbf{0.5648} & \textbf{0.7225} & \textbf{0.1445} & +60.16\% & 136M \\
& $E5_{base}$-v2 + SUTime & 0.3064 & 0.3378 & 0.5662 & 0.1132 & -8.09\% & 109M \\
& $BGE_{large}$-v1.5 & 0.2955 & 0.3269 & 0.5465 & 0.1093 & - & 335M \\
& $BGE_{large}$-v1.5 + TEMPS & 0.5298 & 0.5566 & 0.7070 & 0.1414 & +79.29\% & 362M \\
& $BGE_{large}$-v1.5 + SUTime & 0.2734 & 0.3035 & 0.5268 & 0.1054 & -7.49\% & 335M \\
& Mistral & 0.4063 & 0.4455 & 0.6549 & 0.1310 & - & 7111M \\
& Mistral + TEMPS & \textbf{0.5360} & 0.5622 & 0.7113 & 0.1423 & +31.92\% & 7138M \\
& Mistral + SUTime & 0.3520 & 0.3942 & 0.6254 & 0.1251 & -13.34\% & 7111M \\

\midrule

\multirow{15}{*}{\texttt{TempReason ($\alpha=0.8$)}}
& BM25 & 0.3778 & 0.3512 & 0.5100 & 0.1143 & - & - \\
& SUTime & 0.3188 & 0.2781 & 0.4296 & 0.0976 & - & - \\
& TEMPS w/o Semantic Model & 0.6193 & 0.6285 & 0.7755 & 0.1730 & - & 27M \\
& $MPNet_{base}$-v2 & 0.4957 & 0.5420 & 0.7936 & 0.1724 & - & 109M \\
& $MPNet_{base}$-v2 + TEMPS & 0.6202 & 0.6283 & 0.7732 & 0.1724 & +25.13\% & 136M \\
& $MPNet_{base}$-v2 + SUTime & 0.6466 & 0.6683 & 0.8336 & 0.1823 & +30.45\% & 109M \\
& $E5_{base}$-v2 & 0.6208 & 0.6604 & 0.8587 & 0.1893 & - & 109M \\
& $E5_{base}$-v2 + TEMPS & 0.6278 & 0.6364 & 0.7798 & 0.1738 & +1.13\% & 136M \\
& $E5_{base}$-v2 + SUTime & 0.6592 & 0.6916 & 0.8642 & 0.1903 & +6.20\% & 109M \\
& $BGE_{large}$-v1.5 & 0.5937 & 0.6327 & 0.8413 & 0.1851 & - & 335M \\
& $BGE_{large}$-v1.5 + TEMPS & 0.6286 & 0.6364 & 0.7781 & 0.1735 & +5.88\% & 362M \\
& $BGE_{large}$-v1.5 + SUTime & 0.6532 & 0.6818 & 0.8512 & 0.1874 & +10.03\% & 335M \\
& Mistral & 0.6191 & 0.6731 & 0.8967 & 0.1959 & - & 7111M \\
& Mistral + TEMPS & 0.6320 & 0.6429 & 0.7887 & 0.1756 & +2.07\% & 7138M \\
& Mistral + SUTime & \textbf{0.6901} & \textbf{0.7297} & \textbf{0.9056} & \textbf{0.1981} & +11.45\% & 7111M \\

\midrule

\multirow{15}{*}{\texttt{TS-Retriever ($\alpha=0.3$)}}
& BM25 & 0.0359 & 0.0100 & 0.0133 & 0.0090 & - & - \\
& SUTime & 0.3109 & 0.1915 & 0.2347 & 0.1329 & - & - \\
& TEMPS w/o Semantic Model & 0.1921 & 0.0985 & 0.1111 & 0.0717 & - & 27M \\
& $MPNet_{base}$-v2 & 0.3902 & 0.2396 & 0.2377 & 0.1645 & - & 109M \\
& $MPNet_{base}$-v2 + TEMPS & 0.5596 & 0.3892 & 0.3760 & 0.2687 & +43.39\% & 136M \\
& $MPNet_{base}$-v2 + SUTime & 0.3861 & 0.2375 & 0.2398 & 0.1646 & -1.06\% & 109M \\
& $E5_{base}$-v2 & 0.7880 & 0.6431 & 0.6122 & 0.4238 & - & 109M \\
& $E5_{base}$-v2 + TEMPS & 0.8025 & 0.6726 & 0.6427 & 0.4477 & +1.84\% & 136M \\
& $E5_{base}$-v2 + SUTime & 0.7600 & 0.6228 & 0.6095 & 0.4227 & -3.55\% & 109M \\
& $BGE_{large}$-v1.5 & 0.7151 & 0.5459 & 0.5163 & 0.3600 & - & 335M \\
& $BGE_{large}$-v1.5 + TEMPS & 0.7936 & 0.6463 & 0.6100 & 0.4250 & +10.98\% & 362M \\
& $BGE_{large}$-v1.5 + SUTime & 0.7073 & 0.5415 & 0.5183 & 0.3603 & -1.09\% & 335M \\
& Mistral & 0.7539 & 0.6111 & 0.5856 & 0.4058 & - & 7111M \\
& Mistral + TEMPS & \textbf{0.8240} & \textbf{0.6892} & \textbf{0.6471} & \textbf{0.4508} & +9.29\% & 7138M \\
& Mistral + SUTime & 0.7435 & 0.6041 & 0.5859 & 0.4059 & -1.38\% & 7111M \\

\bottomrule
\end{tabular}
\end{adjustbox}
\end{table*}

To explore our results, we structure our findings into several research questions.

\paragraph{RQ1: Does temporal modeling help with temporal IR?}
Adding TEMPS improves MRR for every base model on every dataset, using the temporal similarity function defined in Section~\ref{sec:system-architecture}, so the temporal component helps both lightweight and large semantic encoders rather than one backbone. The size of the gain and its effect on the remaining metrics differ across benchmarks, and the differences are informative.

On \textbf{TimeQA}, every model augmented with TEMPS shows a large MRR improvement. \textit{Mistral + TEMPS} achieves the highest MRR, while $E5_{base}$-v2 + TEMPS obtains the best NDCG@5, Recall@5, and Precision@5. TimeQA probes sensitivity to \textit{temporal changes in factual knowledge}, embedding explicit time references (e.g., ``in 2007'', ``before 2010'') whose modification changes the correct answer. The gains indicate that TEMPS enables models to go beyond static retrieval or memorization and to make finer temporal distinctions among otherwise relevant passages.

On \textbf{TempReason}, which combines pure temporal reasoning and event temporal grounding, the picture is mixed and we state it plainly. MRR improves for all four backbones, but substantially only for $MPNet_{base}$-v2 ($+25.1\%$), and the $+1.1\%$ gain for $E5_{base}$-v2 is not significant (Appendix~\ref{appendix:significance}). Recall@5 \emph{falls} for all four backbones (down $0.020$, $0.079$, $0.063$ and $0.108$) and NDCG@5 falls for $E5_{base}$-v2 and Mistral, and the rule-based control is the stronger system here: \textit{Mistral + SUTime} posts the best TempReason numbers in Table~\ref{tab:model_performance} on every metric ($0.6901$ MRR, $0.9056$ Recall@5). At the validation-selected $\alpha=0.8$ the fused score is dominated by the temporal branch, which sharpens the head of the ranking at the cost of the top-5 set; RQ5 explains why this benchmark differs from the other two.

On \textbf{TS-Retriever}, improvements remain consistent even for high-performing baselines. \textit{Mistral + TEMPS} reaches the highest MRR, NDCG@5, Recall@5, and Precision@5 in Table~\ref{tab:model_performance}. TEMPS improves both early precision and broad temporal evidence coverage.

\begin{table*}[ht]
\centering
\caption{Comparison of baselines on TS-Retriever. Starred rows are reported by \citet{wu2024time}; we use the same corpus, candidate pool, preprocessing and protocol, which is why those numbers are reused rather than rerun.}
\label{tab:ts_retriever}
\begin{adjustbox}{width=\textwidth}
\begin{tabular}{lcccccccccccc}
    \toprule
    \multirow{2}{*}{\textbf{Model}} & \multicolumn{6}{c}{\textbf{Recall}} & \multicolumn{6}{c}{\textbf{Precision}} \\
    \cmidrule(lr){2-7} \cmidrule(lr){8-13}
     & R@1 & R@5 & R@10 & R@20 & R@50 & R@100 & P@1 & P@5 & P@10 & P@20 & P@50 & P@100 \\
    \midrule
    Contriever\cite{contriever}* & 4.72 & 16.89 & 28.14 & 43.78 & 68.40 & 85.17 & 17.69 & 13.14 & 11.12 & 8.90 & 5.50 & 3.43 \\
    TAS-B\cite{tas}* & 2.62 & 10.17 & 18.40 & 31.02 & 55.18 & 75.56 & 9.49 & 7.27 & 6.56 & 5.56 & 4.03 & 2.85 \\
    JinaAI\cite{gunther2023jina}* & 12.77 & 35.58 & 50.10 & 65.97 & 84.15 & 94.46 & 38.47 & 24.68 & 18.19 & 12.38 & 6.57 & 3.77 \\
    OpenAI* & 9.50 & 29.56 & 40.93 & 55.33 & 77.53 & 92.84 & 29.93 & 20.38 & 14.75 & 10.31 & 6.01 & 3.68 \\
    \midrule
    TSContriever* & \underline{19.92} & \underline{52.70} & \underline{69.13} & \underline{81.52} & \underline{91.53} & 94.57 & \underline{58.63} & \underline{36.49} & \underline{25.34} & \underline{15.72} & \underline{7.30} & 3.81 \\
    TSContriever*$_{\text{only-q}}$ & 11.62 & 38.96 & 56.38 & 74.44 & 89.88 & 94.82 & 41.27 & 29.15 & 21.60 & 14.54 & 7.18 & 3.81 \\
    + query-router* & 10.94 & 36.71 & 52.95 & 71.01 & 88.94 & \underline{95.11} & 39.49 & 27.67 & 20.40 & 14.02 & 7.15 & \underline{3.83} \\
    \midrule
    $E5_{\text{base}}$-v2 + TEMPS ($\alpha=0.3$)
& \textbf{25.39} & \textbf{64.19} & \textbf{80.36}
& \textbf{90.92} & \textbf{97.08} & \textbf{98.95}
& \textbf{69.14} & \textbf{44.59} & \textbf{30.23}
& \textbf{17.83} & \textbf{7.77} & \textbf{3.98} \\
    \bottomrule
\end{tabular}
\end{adjustbox}
\end{table*}

Table~\ref{tab:ts_retriever} compares $E5_{base}$-v2 + TEMPS with the state-of-the-art \textit{TSContriever} models. Our setup matches \citet{wu2024time} exactly, in corpus, candidate pool, preprocessing, and evaluation protocol, which is why their published numbers are reused rather than rerun. This configuration surpasses \textit{TSContriever} not only at deeper ranks but also at rank 1, improving R@1 from 19.92 to 25.39 and P@1 from 58.63 to 69.14. It also maintains stronger recall across the ranking window, reaching R@100 of 98.95 compared to 94.57 for \textit{TSContriever}, while improving deeper precision such as P@20 (17.83 vs. 15.72).

To address the fact that \textit{TSContriever} is purely temporal and trained on TS-Retriever, Wu et al.~\cite{wu2024time} proposed a hybrid query-router architecture that sends time-sensitive queries to \textit{TSContriever} and other queries to the standard semantic \textit{Contriever}. Our unified approach outperforms this composite system on every metric, suggesting that TEMPS can combine temporal sensitivity with semantic matching without requiring a separate routing mechanism.

These results demonstrate that TEMPS addresses subtle weaknesses in handling \textit{semantically similar but temporally mismatched distractors}.

\paragraph{RQ2: Do temporal retrieval gains improve downstream RAG?}

\begin{table}[t]
\centering
\caption{Downstream RAG performance on the held-out TimeQA test split, using Qwen2.5-7B-Instruct as the generator and the selected $\alpha=0.6$.}
\label{tab:rag_timeqa}
\begin{adjustbox}{width=\linewidth}
\begin{tabular}{lcccc}
    \toprule
    \textbf{Retriever} & \textbf{$\alpha$} & \textbf{EM} & \textbf{F1} & \textbf{Containment} \\
    \midrule
    TEMPS w/o Semantic Model & -- & 0.3521 & 0.4340 & 0.4085 \\
    $E5_{base}$-v2 & -- & 0.2915 & 0.3644 & 0.3352 \\
    $E5_{base}$-v2 + TEMPS & 0.60 & \textbf{0.3845} & \textbf{0.4665} & \textbf{0.4423} \\
    $BGE_{large}$-v1.5 & -- & 0.2648 & 0.3272 & 0.3099 \\
    $BGE_{large}$-v1.5 + TEMPS & 0.60 & 0.3746 & 0.4563 & 0.4324 \\
    $MPNet_{base}$-v2 & -- & 0.2155 & 0.2653 & 0.2493 \\
    $MPNet_{base}$-v2 + TEMPS & 0.60 & 0.3718 & 0.4507 & 0.4254 \\
    Mistral & -- & 0.3028 & 0.3799 & 0.3592 \\
    Mistral + TEMPS & 0.60 & 0.3761 & 0.4620 & 0.4394 \\
    \bottomrule
\end{tabular}
\end{adjustbox}
\end{table}

Table~\ref{tab:rag_timeqa} evaluates whether the retrieval improvements translate into end-to-end answer quality. Using the same TimeQA queries and Qwen2.5-7B-Instruct generator, every +TEMPS retriever improves EM, F1, and containment over its unaugmented counterpart. The strongest overall RAG performance comes from $E5_{base}$-v2 + TEMPS, with similar gains across the other retrievers.

These RAG results provide a downstream validation of the TTS objective: better temporal alignment in retrieval leads to better generated answers, not only better ranking scores. They also address the practical concern that improvements at deeper retrieval ranks may not affect top-sensitive pipelines. Although the generator sees only the top-5 retrieved passages, temporalized retrievers consistently produce more correct final answers, indicating that TEMPS improves the evidence available to the generator in the part of the ranking that directly matters for RAG. This is our only end-to-end evaluation: a second answer-level task needs both genuine temporal retrieval constraints and trustworthy answer-level ground truth under a comparable protocol, and no benchmark we examined supplied both.

\paragraph{RQ3: Is it better to increase the size of the model or to better model time in temporal IR?}
Our results show that correctly modeling temporal information yields larger gains than simply scaling model size. Across all three benchmarks, $E5_{base}$-v2 + TEMPS obtains a higher MRR than the unaugmented 7B-parameter \textit{Mistral} model despite having around 50 times fewer parameters, though on TempReason the margin ($0.6278$ vs.\ $0.6191$) is too small to carry weight and the 7B model remains well ahead on Recall@5. Scaling remains useful when the temporal module is also added, as \textit{Mistral + TEMPS} obtains the best MRR on all three datasets. However, the unaugmented 7B model is consistently weaker than much smaller temporalized encoders. These results highlight that temporal modeling can outweigh raw capacity increases, especially on tasks with strong temporal dependencies, which is important for resource-constrained retrieval systems.

\paragraph{RQ4: What is the importance of time in the model’s scoring function?}
An $\alpha$ sweep (Appendix~\ref{sec:appendix-alpha-sensitivity}) confirms performance peaks at non-zero values across all datasets, proving temporal information is critical. We calibrate $\alpha$ per benchmark on validation MRR to balance semantic relevance and temporal compatibility, as a single global value would underperform on diverse tasks. For Table~\ref{tab:model_performance}, this gives: TimeQA ($\alpha=0.6$), TempReason ($\alpha=0.8$), and TS-Retriever ($\alpha=0.3$). The sweep endpoints are the two ablations in the same table: $\alpha=0$ is each semantic-only baseline row and $\alpha=1$ is \textit{TEMPS w/o Semantic Model}. TimeQA and TempReason benefit from higher temporal weights due to their focus on evolving facts and date arithmetic. Conversely, TS-Retriever requires a lower weight to prevent time from suppressing the semantic signals needed to distinguish topically close distractors.

\paragraph{RQ5: Does the learned encoder beat rule-based grounding plus the same KL score?}
On TimeQA and TS-Retriever, clearly yes. Substituting SUTime for the learned encoder does not merely lose the gain, it falls below the semantic baseline: MRR drops by $7.5$--$13.3\%$ across the four TimeQA backbones and by $1.1$--$3.6\%$ on TS-Retriever. Because the control holds the fusion, normalization, and $\alpha$ fixed, this isolates the temporal representation as the source of the improvement.

On TempReason the control wins instead, by $0.025$ to $0.058$ MRR, and the reason is benchmark composition rather than a defect in the fusion. On the $4{,}306$ held-out queries that carry an explicit year, exact-year and query-interval regex oracles each lift MRR by only $0.009$ over the semantic baseline, while our temporal head alone reaches $0.8695$, above both, before any semantic fusion; in $48\%$ of those queries the gold passage carries no year from the query interval at all (Appendix~\ref{sec:appendix-oracle}). The gains therefore do not reduce to year matching. But TempReason queries are regular enough that a tuned rule-based pipeline stays competitive end to end, and the learned encoder buys little where a regex already suffices.

\paragraph{RQ6: Can temporal modeling be integrated without huge computational cost?}
TEMPS adds only a lightweight temporal embedding module to an existing model, increasing parameter count by approximately 27M in our experiments. Even when applied to small embedding models like $E5_{base}$-v2, it delivers performance competitive with, and often exceeding, the unaugmented 7B-parameter Mistral model. Furthermore, TEMPS can be attached to different semantic embedding models without retraining the semantic component, making it a modular addition to existing retrieval pipelines. This makes the approach suitable for real-world systems where both latency and memory footprint are critical constraints. Finally, Appendix~\ref{sec:appendix-qualitative} pairs the semantic-only, temporal-only, and +TEMPS rankings per TimeQA query and reports where the gold passage moves, in a setting where candidates come from a single article and differ mainly in temporal scope.

\section{Conclusion}
\label{sec:conclusion}

While large language models and dense retrieval methods have advanced the field of information retrieval, they often fall short in addressing temporal reasoning, an essential factor for accurately interpreting time-sensitive user queries. To tackle this limitation, we proposed a novel framework called \textbf{TEMPS} based on the Temporal Textual Similarity task, leveraging weak supervision and synthetic data to train a dedicated embedding model that captures temporal information in documents. Our evaluations show MRR improvements over existing baselines for every semantic backbone tested. A rule-based control that swaps SUTime grounding into the same fusion falls below the semantic baseline on TimeQA and TS-Retriever, locating the gain in the learned representation; on TempReason it is stronger, and TEMPS trades top-5 coverage for a better top-1. The benefit is concentrated where topical and temporal relevance conflict, not on queries a tagger already resolves.

\paragraph{Future Work} While our explicit temporal modeling proves effective, several promising directions remain open. First, tighter integration of semantic and temporal embeddings, via joint training or attention-based fusion, could improve robustness against semantically similar yet temporally incongruent distractors. Second, extending TEMPS to \textit{implicit} temporal references and event-based reasoning would broaden its applicability to real-world narrative and conversational domains. Third, scaling data generation to diverse large-scale open corpora could enhance coverage of rare temporal expressions. Finally, we aim to explore multilingual retrieval and lightweight modules for low-latency, resource-constrained environments.

\section*{Limitations}
Our evaluation focuses on benchmarks with explicit temporal annotations and reliable timestamps, enabling controlled comparison but not covering all real-world variability in temporal language or metadata quality. All experiments are conducted in English; we have not evaluated TEMPS on other languages, and its effectiveness in multilingual or cross-lingual settings remains an open question. Our temporal supervision is built from synthetic templates and a rule-based, HeidelTime-tagged resource, and it grounds anchored, resolvable temporal expressions; implicit or highly ambiguous references and event–event relations (e.g., TempReason Level 3) are outside our current scope, and the weak supervision inherits any noise from the underlying tagger. The temporal–semantic weight $\alpha$ is selected per benchmark on a held-out validation split from MRR alone, which keeps the protocol uniform but accepts a coverage cost: on TempReason, where validation MRR peaks at $\alpha=0.8$, adding TEMPS lowers Recall@5 for all four backbones and NDCG@5 for two of them, and a rule-based SUTime control outperforms it on every metric. We report this rather than tune around it; the benefit of the learned encoder is concentrated where topical and temporal relevance conflict, not on queries an off-the-shelf tagger already resolves. Reported results come from a single training run (seed 42), with statistical significance measured across queries rather than across training seeds. TEMPS is also model-agnostic and thus inherits the biases and factual gaps of the semantic backbone it augments, a trade-off we accept to ensure modularity and ease of integration. The retrieval unit is a passage of up to 512 tokens, and attention pooling collapses however many temporal expressions it contains into a single Gaussian, which is a lossy summary for long or temporally heterogeneous passages; passages with no temporal expression are still encoded at inference but are held out of temporal training, so their Gaussians are uncalibrated and the semantic branch is what ranks them. Our downstream evidence rests on a single end-to-end task (TimeQA RAG with Qwen2.5-7B-Instruct), because we found no second benchmark supplying both genuine temporal retrieval constraints and trustworthy answer-level ground truth under a comparable protocol. Finally, while the added temporal encoding introduces a small computational overhead, it remains far lower than the cost of scaling to much larger models for comparable gains.
\section*{Ethical Considerations}
TEMPS is foundational retrieval research and is not tied to a specific deployment. Because it is model-agnostic, it inherits any biases or factual gaps of the semantic backbone it augments; in time-sensitive, high-stakes domains such as clinical or legal search, temporally miscalibrated retrieval could surface misleading evidence, and downstream applications should therefore validate temporal alignment before relying on it. We use only synthetic and publicly available research data and release our artifacts for research use only.

\section*{Acknowledgments}
This project was provided with computing and storage resources by GENCI at IDRIS thanks to the grant 2025-105442 on the supercomputer Jean Zay's A100 partition.

\bibliography{references}

\appendix

\section{Dataset Construction Details}
\label{sec:appendix-dataset}
\subsection{Training Dataset Construction}

\begin{figure*}
    \centering
    \includegraphics[width=\linewidth]{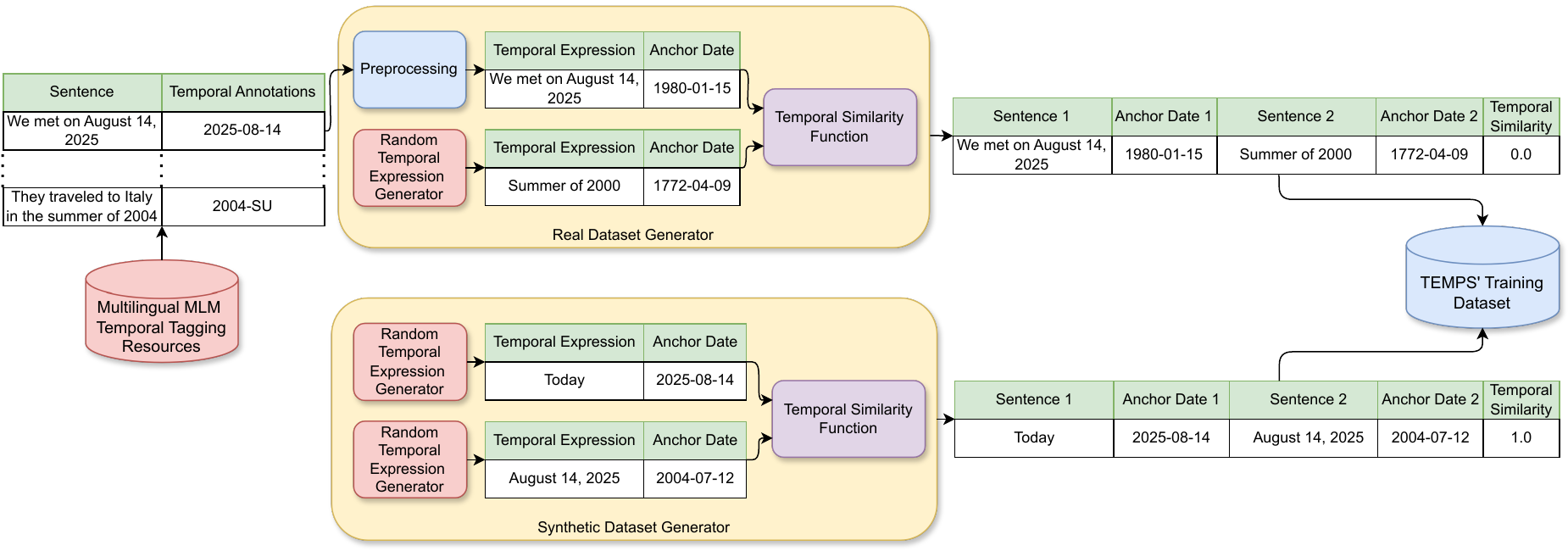}
    \caption{TEMPS' Training Dataset Creation Pipeline}
    \Description[Training Dataset Creation Pipeline]{Training Dataset Creation Pipeline}
    \label{fig:current-data-pipeline-architecture}
\end{figure*}

Figure~\ref{fig:current-data-pipeline-architecture} describes our training dataset generation pipeline. The construction is based on two main components: generating synthetic data based on predefined patterns and adapting an existing dataset for our task. The pipeline produces a dataset composed of tuples $(s_1, d_1, s_2, d_2, ts)$, where $s_1$ and $s_2$ are texts containing temporal expressions, $d_1$ and $d_2$ are anchor dates at a daily granularity (YYYY-MM-DD), and $ts$ is the temporal label between $(s_1, d_1)$ and $(s_2, d_2)$ as defined by our temporal similarity function (see Section~\ref{sec:tsf}).

\subsubsection{Temporal Similarity Function}
\label{sec:tsf}

\paragraph{Setup}
Given two TimeML annotations grounded as closed integer day intervals $I_q = [a_q, b_q]$ and $I_d = [a_d, b_d]$, where a single date is the degenerate case $a=b$ and year or month granularities map to the full corresponding range, we define a function $\mathrm{TSF}(I_q \!\to\! I_d) \in \mathbb{R}_{\geq 0}$ that quantifies how well the temporal extent of the document interval $I_d$ accommodates that of the query interval $I_q$. The function serves as a soft supervision signal for contrastive training of a temporal embedding model whose inference-time similarity adopts the asymmetric Gaussian-KL scoring used in prior distributional embedding work~\cite{vilnis2014word, he2015learning, yoda2023sentence}.

We require three properties. First, the function should be \textbf{asymmetric}, in order to encode directional relations from Allen's interval algebra~\citep{allen1983maintaining}, such as \textit{during} ($I_q$ is contained in $I_d$) and \textit{contains}, as distinct from their inverses. Second, it should be \textbf{non-degenerate on disjoint pairs}, ranking far-apart pairs differently from near-disjoint pairs rather than collapsing both to zero. Finally, it should be \textbf{continuous and smooth} in the four endpoints, with no discontinuity at the overlap/disjoint boundary.

\paragraph{Construction}
We treat each grounded interval as a uniform probability distribution over its temporal extent, $T_I \sim \mathrm{U}[a, b + 1)$, where the $+1$ maps the discrete inclusive interval to a continuous half-open interval. Two annotations are then compared as distributions rather than as sets.

For analytical tractability and to match the embedding head described in Section~\ref{sec:system-architecture}, we project each uniform distribution onto the Gaussian family by moment matching:
\begin{equation}
\begin{aligned}
p_I &= \mathcal{N}\!\bigl(\mu_I, \sigma_I^2\bigr), \\
\mu_I &= \frac{a + b + 1}{2}, \\
\sigma_I^2 &= \frac{(b - a + 1)^2}{12}.
\end{aligned}
\end{equation}
These are the exact first and second central moments of $T_I$. The projection preserves temporal location through the mean and the uncertainty implied by granularity through the variance, while admitting closed-form divergences. Both parameters depend only on $(a,b)$; no global constants enter the construction.

\paragraph{Why moment matching, and not the uniforms themselves}
\label{sec:moment-matching}
The obvious alternative is to score the interval uniforms directly, without any
Gaussian. It does not work, because the KL divergence between two uniforms is
degenerate as a supervision signal. For $T_{I_q}\sim\mathrm{U}[a_q,b_q+1)$ and
$T_{I_d}\sim\mathrm{U}[a_d,b_d+1)$,
\[
D_{\mathrm{KL}}(T_{I_q}\Vert T_{I_d}) =
\begin{cases}
\log\dfrac{b_d-a_d+1}{b_q-a_q+1}, & I_q \subseteq I_d,\\[6pt]
+\infty, & \text{otherwise,}
\end{cases}
\]
because the support of $T_{I_q}$ must lie inside the support of $T_{I_d}$ for the
density ratio to stay finite. The divergence is therefore finite only under full
containment, and infinite the moment any part of $I_q$ falls outside $I_d$. A
one-day overhang and a thirty-year separation receive the same value. Worse, the
finite branch depends only on the ratio of widths and not at all on where the
intervals sit, so two containments with very different alignments are
indistinguishable.

Moment matching keeps what is useful in this and discards what is not. It
preserves the direction, since the variance ratio $\sigma_q^2/(2\sigma_d^2)$ is
still small when the query is sharper than the document and large in reverse, so
\textit{during} and \textit{contains} remain distinguishable. It replaces the
single infinity with the term $(\mu_q-\mu_d)^2/(2\sigma_d^2)$, which grows
quadratically with separation and orders the disjoint cases the uniform KL
collapses. And it is smooth across the containment boundary, so the label
function has no discontinuity for the model to fit around. The cost is that the
Gaussian assigns non-zero density outside the interval; we accept it because the
label is only ever used to induce an ordering, and the ordering is what the
CoSENT objective consumes.

\paragraph{Definition}
We define the temporal similarity function as the asymmetric Kullback-Leibler divergence from the query distribution to the document distribution:
\begin{equation}
\begin{aligned}
\mathrm{TSF}(I_q \!\to\! I_d)
&:= D_{\mathrm{KL}}\!\bigl(p_{I_q} \,\Vert\, p_{I_d}\bigr) \\
&= \frac{1}{2}\log\!\frac{\sigma_d^2}{\sigma_q^2} \\
&\quad
+ \frac{\sigma_q^2 + (\mu_q - \mu_d)^2}{2\,\sigma_d^2}
- \frac{1}{2}.
\end{aligned}
\end{equation}
Smaller values indicate that $I_q$ is more temporally consistent with $I_d$. The function is non-negative and is zero iff $p_{I_q}=p_{I_d}$, i.e., iff $I_q$ and $I_d$ are identical intervals.

When a bounded similarity in $(0,1]$ is required, such as for losses that expect similarity rather than distance, we use the monotone-decreasing transform
\[
\sigma(I_q \!\to\! I_d) = \frac{1}{1 + \mathrm{TSF}(I_q \!\to\! I_d)},
\]
mirroring the form of the model's inference-time similarity head.

\paragraph{Properties}
\textbf{Asymmetry.} When $\sigma_q \neq \sigma_d$, the function is asymmetric in its arguments. The variance term $\sigma_q^2/(2\sigma_d^2)$ is small when the query is sharper than the document and large in the reverse direction. For a day $d$ inside its containing year $Y$, an instance of Allen's \textit{during} relation, $\mathrm{TSF}(d \!\to\! Y)$ is small because the day is well explained by the year, whereas $\mathrm{TSF}(Y \!\to\! d)$ is large because the year's broad distribution is poorly explained by a single day. This is the kind of inclusion asymmetry that Gaussian-KL embeddings have previously modeled in lexical, graph, and knowledge-graph settings~\citep{vilnis2014word, he2015learning, bojchevski2018deep}.

\textbf{Non-degeneracy.} The mean-squared term $(\mu_q-\mu_d)^2/(2\sigma_d^2)$ is unbounded above and grows quadratically with temporal separation. Distant disjoint pairs therefore receive larger labels than near-disjoint pairs, providing a well-ordered ranking signal across the full input space.

\textbf{Smoothness.} $\mathrm{TSF}$ is $C^\infty$ in $(\mu_q,\sigma_q,\mu_d,\sigma_d)$ on $\sigma_q,\sigma_d>0$, and therefore $C^\infty$ in the four endpoint parameters. The overlap/disjoint boundary is invisible to the function.

\textbf{Granularity invariance.} The same closed form applies regardless of whether $I_q$ and $I_d$ have matched granularity. Coarse-fine pairs, such as a year and a day, and same-granularity pairs, such as two years, are handled uniformly, with the resulting asymmetry direction aligned with the inclusion relation. No special-case branching is required.

\textbf{Compatibility with the embedding head.} Our model emits a Gaussian $\mathcal{N}(\hat{\mu}, \hat{\sigma}^2)$ per text-date input and scores candidate pairs with $1/(1+D_{\mathrm{KL}}(\hat{p}_q \Vert \hat{p}_d))$ (Section~\ref{sec:system-architecture}), mirroring the bounded inverse-KL score used by GaussCSE~\citep{yoda2023sentence}. With $\mathrm{TSF}$ as the label function, training reduces to same-family distillation: the model learns to map anchored textual surface forms to Gaussian parameters that reproduce the closed-form temporal KL of the underlying intervals. The temporal-specific contribution is the interval grounding and moment-matched supervision, not the Gaussian-KL score itself.

\begin{table}[t]
\centering
\small
\begin{tabular}{lrr}
\toprule
Pair & $q \!\to\! d$ & $d \!\to\! q$ \\
\midrule
Same interval & $0.0$ & $0.0$ \\
Year vs.\ day-in-middle & $6.8\!\times\!10^4$ & $5.4$ \\
Year vs.\ day-at-boundary & $2.7\!\times\!10^5$ & $6.9$ \\
Decade vs.\ year-inside & $48.8$ & $1.8$ \\
Adjacent years & $6.0$ & $6.0$ \\
Far-disjoint years (1990, 2025) & $	7.36\!\times\!10^3$& $	7.36\!\times\!10^3$\\
1-day overlap & $4.86$ & $4.86$ \\
1-day gap & $6.00$ & $6.00$ \\
\bottomrule
\end{tabular}
\caption{$\mathrm{TSF}$ values on canonical interval pairs, in both directions. Containment relations exhibit the targeted asymmetry, same-width pairs are symmetric, and far-disjoint pairs receive labels distinct from near-disjoint pairs.}
\label{tab:tsf-examples}
\end{table}

\paragraph{Numerical illustration}
Table~\ref{tab:tsf-examples} reports $\mathrm{TSF}$ on canonical pairs. For a day in the middle of a containing year, $\mathrm{TSF}(\text{day}\!\to\!\text{year})=5.4$ while $\mathrm{TSF}(\text{year}\!\to\!\text{day})=6.8\times 10^4$. For two adjacent same-width years, the function is symmetric, with both directions equal to $6.0$. For far-disjoint years it grows quadratically with separation, reaching $1.56\times 10^6$ for 1990 vs.\ 2025. The transition from overlapping to gapped intervals is smooth: a one-day overlap gives $4.86$ and a one-day gap gives $6.00$, unlike IoU-based schemes that exhibit a discontinuity at the same boundary.

\subsubsection{Data Sources}

\paragraph{Synthetic Dataset}
To enable scalable and diverse temporal understanding, we construct a synthetic dataset spanning the years 1000 to 2030. TimeML annotations are randomly generated and converted into natural language using predefined templates (Table~\ref{tab:templates}). Each expression is paired with a randomly assigned anchor date (daily granularity) to resolve context-dependent expressions.

\begin{table}[h]
\centering
\caption{Sample TimeML Annotations with Variable Offsets}
\label{tab:templates}
\begin{tabularx}{\columnwidth}{|p{3.5cm}|X|}
\hline
\textbf{TimeML Annotation} & \textbf{Text Template} \\
\hline
OFFSET P\{n\}D & "\{n\} days ago/later" \\
\hline
OFFSET P\{n\}W & "\{n\} weeks ago/later" \\
\hline
OFFSET P\{n\}M & "\{n\} months ago/later" \\
\hline
OFFSET P\{n\}Y & "\{n\} years ago/later" \\
\hline
THIS P\{n\}D & "these \{n\} days" \\
\hline
THIS P\{n\}W & "these \{n\} weeks" \\
\hline
THIS P\{n\}M & "these \{n\} months" \\
\hline
THIS P\{n\}Y & "these \{n\} years" \\
\hline
\{date1\},\{date2\} & "spanning from \{date1\} to \{date2\}" \\
\hline
\{date1\},\{date2\} & "between \{date1\} and \{date2\}" \\
\hline
\end{tabularx}
\end{table}

\paragraph{Multilingual MLM Temporal Tagging Dataset}
We incorporate the English portion of the Multilingual MLM Temporal Tagging Resources dataset~\cite{lange2022multilingual}. This dataset provides text from Wikipedia with temporal expressions automatically annotated using a modified HeidelTime~\cite{strotgen2010heideltime} tagger. This serves as a source of weakly supervised, real-world linguistic patterns to complement our synthetic data.

\subsubsection{Training Instance Generation}

We apply a unified processing pipeline to both data sources to create a contrastive learning dataset. For every temporal expression $t$ found in either source, we generate one positive sample ($s_0$) and four negative samples ($s_1...s_4$).

The positive sample is created by slightly modifying the boundaries of $t$ while
maintaining a bounded temporal similarity $>0.9$, and generating a corresponding
textual representation. Negative samples are drawn at random, with random anchor
dates, while the positive sample retains the original or logically derived
anchor. Random negatives may be temporally easy, but this concerns training only:
the evaluation candidate pools are fixed by the benchmarks and are never
resampled, so the reported retrieval results do not depend on this choice. Mining
hard temporal negatives, where the negative is a near-miss in time rather than a
random draw, is a natural extension we leave to future work.

The final dataset consists of 60M tuples $(s_1, d_1, s_2, d_2, ts)$, obtained by concatenating 55M synthetic tuples and 5M real-world tuples. Here, $s_1$ and $s_2$ are temporal expression strings, $d_1$ and $d_2$ are their respective anchor dates (\texttt{YYYY-MM-DD}), and $ts \in (0, 1]$ is a continuous temporal similarity score, not a discrete class: it is the bounded value $ts = 1 / \bigl(1 + \mathrm{TSF}(I_1 \rightarrow I_2)\bigr)$ obtained by applying the monotone-decreasing transform of Appendix~\ref{sec:tsf} to the directional temporal similarity function, with $ts = 1$ iff the two grounded intervals are identical.

\section{Temporal Model Details}
\label{sec:appendix-model-details}

\paragraph{Input Encoding.}
The temporal branch operates on a sentence and an anchor date. The sentence is processed by a frozen all-MiniLM-L6-v2 backbone to produce contextualized hidden states $\mathbf{H}$. The anchor date is encoded using a sinusoidal positional encoder following~\cite{jia2021complex}. Let $k \in \mathbb{N}$ be the number of days between the input date and 1 January 1000. The encoder is defined as:
\[
\begin{aligned}
PE_{2i}(t) &= \sin\left(\frac{k}{n^{2i/d}}\right), \\
PE_{2i+1}(t) &= \cos\left(\frac{k}{n^{2i/d}}\right),
\end{aligned}
\]
where $d$ is the output embedding dimension, set to $32$ in our experiments, $i \in [0,d/2]$, and $n$ is a frequency scaling factor.

\paragraph{Gaussian Embedding Head.}
The hidden states are aggregated with attention pooling to emphasize temporal tokens:
\[
\mathbf{e}_{\text{sent}} = \mathrm{AttentionPooling}(\mathbf{H}).
\]
The pooled sentence vector is concatenated with the date representation $\mathbf{p}_{\text{date}}$ and passed through a temporal learning network:
\[
\mathbf{z} = \mathrm{TemporalLearning}(\mathbf{e}_{\text{sent}} \oplus \mathbf{p}_{\text{date}}).
\]
Two output heads then estimate the Gaussian parameters:
\[
\mu = \mathrm{MeanNN}(\mathbf{z}), \qquad
\sigma = \mathrm{VarianceNN}(\mathbf{z}).
\]
We use only the diagonal elements of the covariance matrix for computational efficiency~\cite{vilnis2014word, yoda2023sentence}. Gaussian embeddings are useful because the mean represents temporal location and the variance represents uncertainty or granularity, enabling asymmetric inclusion behavior through KL divergence.

\paragraph{Training Objective.}
Let $(s^i_1,d^i_1,s^i_2,d^i_2,ts_i)$ be a training point. The model produces Gaussian embeddings for both anchored expressions, and their bounded temporal score is:

\begin{equation}
\begin{aligned}
    s_i &= \frac{1}{1+D_{\mathrm{KL}}\!\big(p^i_1 \,\Vert\, p^i_2\big)}, \\[4pt]
    p^i_j &= \mathcal{N}\!\big(\mu^i_j,\,\mathrm{diag}((\sigma^i_j)^2)\big).
\end{aligned}
\end{equation}

We adapt the CoSENT loss~\cite{10.1109/TASLP.2024.3402087} to enforce the ordering induced by the ground-truth temporal labels:
\begin{equation}
    \mathcal{L} = \log \left( 1 + \sum_{i,j :\, ts_i < ts_j} \exp \left( \lambda \cdot (s_i - s_j) \right) \right),
    \label{eq:cosent_loss}
\end{equation}
where $\lambda=20$ in our experiments.

\paragraph{Training Configuration.} The model was trained for three epochs over the full training corpus on a single HPC \texttt{gpu\_p5} node with exclusive allocation. The node provides two AMD EPYC 7543 CPUs, 512\,GiB of DDR4 memory, and eight NVIDIA A100-SXM4 GPUs with 80\,GB of HBM2 memory each. Training used PyTorch's distributed launcher (\texttt{torchrun}, one process per GPU) with a global batch size of 2{,}048, a learning rate of $3 \times 10^{-4}$, weight decay of $0.01$, a linear warmup over the first 5\% of optimization steps, a maximum input length of 512 tokens, 16-bit floating-point precision, and random seed 42. Intermediate validation is performed every 500 optimization steps.

\section{Experimental Details}
\label{sec:appendix-experimental-details}

\subsection{Baselines}

To evaluate the effectiveness of our proposed temporal embedding model, we benchmark its performance against a set of baselines stratified by model size. This allows us to assess whether simply scaling up model parameters captures temporal nuances or if specialized architecture is required regardless of size.

We categorize our dense retrieval baselines into three distinct tiers based on parameter count:

\begin{itemize}
    \item \textbf{Base-Scale Models:} We employ \textbf{MPNet\textsubscript{base}-v2}\footnote{\texttt{sentence-transformers/all-mpnet-base-v2}}~\cite{mpnet} and \textbf{E5\textsubscript{base}-v2}\footnote{\texttt{intfloat/e5-base-v2}}~\cite{e5}. These models represent the standard for efficient, low-latency deployment. We include them to determine if our method provides significant gains in resource-constrained environments where larger models are impractical.
    
    \item \textbf{Large-Scale Models:} We include \textbf{BGE\textsubscript{large}-v1.5}\footnote{\texttt{BAAI/bge-large-en-v1.5}}~\cite{bge} to represent the upper bound of traditional BERT-based encoder architectures. This baseline tests whether increasing the capacity of pure encoder models is sufficient to resolve complex temporal queries.
    
    \item \textbf{LLM-Based Models:} We use \textbf{Mistral}\footnote{\texttt{Salesforce/SFR-Embedding-Mistral}}, a state-of-the-art retrieval model finetuned from Mistral-7B. By comparing against a model orders of magnitude larger, we investigate whether the emergent reasoning capabilities of Large Language Models render explicit temporal modeling obsolete.

    \item \textbf{State-of-the-Art Temporal Baseline}: We compare against TSContriever~\cite{wu2024time}, a dense retrieval model explicitly optimized for time-sensitive questioning. We also include its variant, TSContriever + Query Router, which selectively employs temporal reasoning modules. Including these ensures we benchmark TEMPS against the current specialized state-of-the-art, not just general-purpose embeddings.
\end{itemize}

In addition to these dense retrievers, we include \textbf{BM25} (lexical baseline) and \textbf{SUTime}~\cite{chang2012sutime} (rule-based baseline). Finally, we compare all dense models against their augmented versions (e.g., \textit{Mistral + TEMPS}) to isolate the specific performance contributions of our temporal module.

\paragraph{SUTime + KL rule-based control.}
The \textit{+ SUTime} rows in Table~\ref{tab:model_performance} are an ablation,
not a separate system. They answer whether the learned encoder is needed at all,
given that the supervision already grounds intervals in closed form. The control
changes exactly one component of the +TEMPS pipeline: SUTime grounds the temporal
expressions in the query and the candidate to intervals, those intervals are
mapped to Gaussians by the same moment matching (Eq.~\ref{eq:moment-match}), and
the resulting temporal score is fused with the semantic score under the same
inverse-KL form, the same per-query min-max normalization, and the same
validation-selected $\alpha$. Everything downstream of the temporal
representation is held fixed, so the difference between a \textit{+ TEMPS} row
and its \textit{+ SUTime} row measures the representation and nothing else.

\subsection{Benchmarks and Metrics}

We evaluate our system using three temporal benchmarks: \textbf{TimeQA}~\cite{chen2021dataset} assesses the capability to understand the time scope of evolving facts within long documents, specifically evaluating \textbf{temporal understanding} of anchored and resolvable time mentions. \textbf{TempReason}~\cite{tan2023towards} evaluates specific reasoning levels, combining abstract logic rules (Level 1) and grounding events to valid time ranges (Level 2). We omit Level 3 from this study, as it focuses on event-event relations, whereas our research targets temporal relations covered by the first two levels. \textbf{TS-Retriever}~\cite{wu2024time} focuses on the retrieval stage by testing the distinction of semantically similar documents based on temporal constraints via \textbf{explicit} (specific years), \textbf{implicit} (vague periods), and \textbf{hybrid} (ranges) query types. Together, these benchmarks provide a comprehensive assessment of factual, retrieval-based, and arithmetic temporal reasoning, sharing the common feature of evaluating time-sensitive retrieval from a \textbf{given set of documents}. We additionally use TimeQA for downstream RAG evaluation by passing the top-5 retrieved passages to Qwen2.5-7B-Instruct~\cite{qwen2024qwen25} and evaluating the generated answer.

\begin{table}[h]
\centering
\begin{tabular}{l r r r}
\toprule
\textbf{Dataset} & \textbf{Total} & \textbf{Val (20\%)} & \textbf{Test (80\%)} \\
\midrule
TimeQA       & 887  & 177   & 710 \\
TS-Retriever & 3244 & 649   & 2595 \\
TempReason   & 9397 & 1879  & 7518 \\
\bottomrule
\end{tabular}
\caption{Evaluation datasets and the random validation/test partition used for
held-out selection of $\alpha$. The validation split is used only to choose one
$\alpha$ per benchmark from MRR; every number reported in the paper is measured
on the test split.}
\label{tab:datasets}
\end{table}

\paragraph{Held-out protocol.}
Each benchmark is partitioned once, at random, into a 20\% validation split and
an 80\% test split (Table~\ref{tab:datasets}). We sweep $\alpha$ on validation
MRR alone and fix the winning value for that benchmark. That value is then applied unchanged to every backbone, every metric,
and every significance test, and the test split is scored once. No $\alpha$ is
tuned per backbone or per metric.

Retrieval effectiveness is evaluated using four ranking metrics: MRR, NDCG@5, Recall@5, and Precision@5. Downstream RAG effectiveness is evaluated separately using exact match (EM), token-level F1, and answer containment, where containment measures whether a normalized gold answer string appears in the generated response.

\subsection{Effect of Alpha Values}
\label{sec:appendix-alpha-sensitivity}

\begin{figure*}[t]
    \centering
    \includegraphics[width=\textwidth]{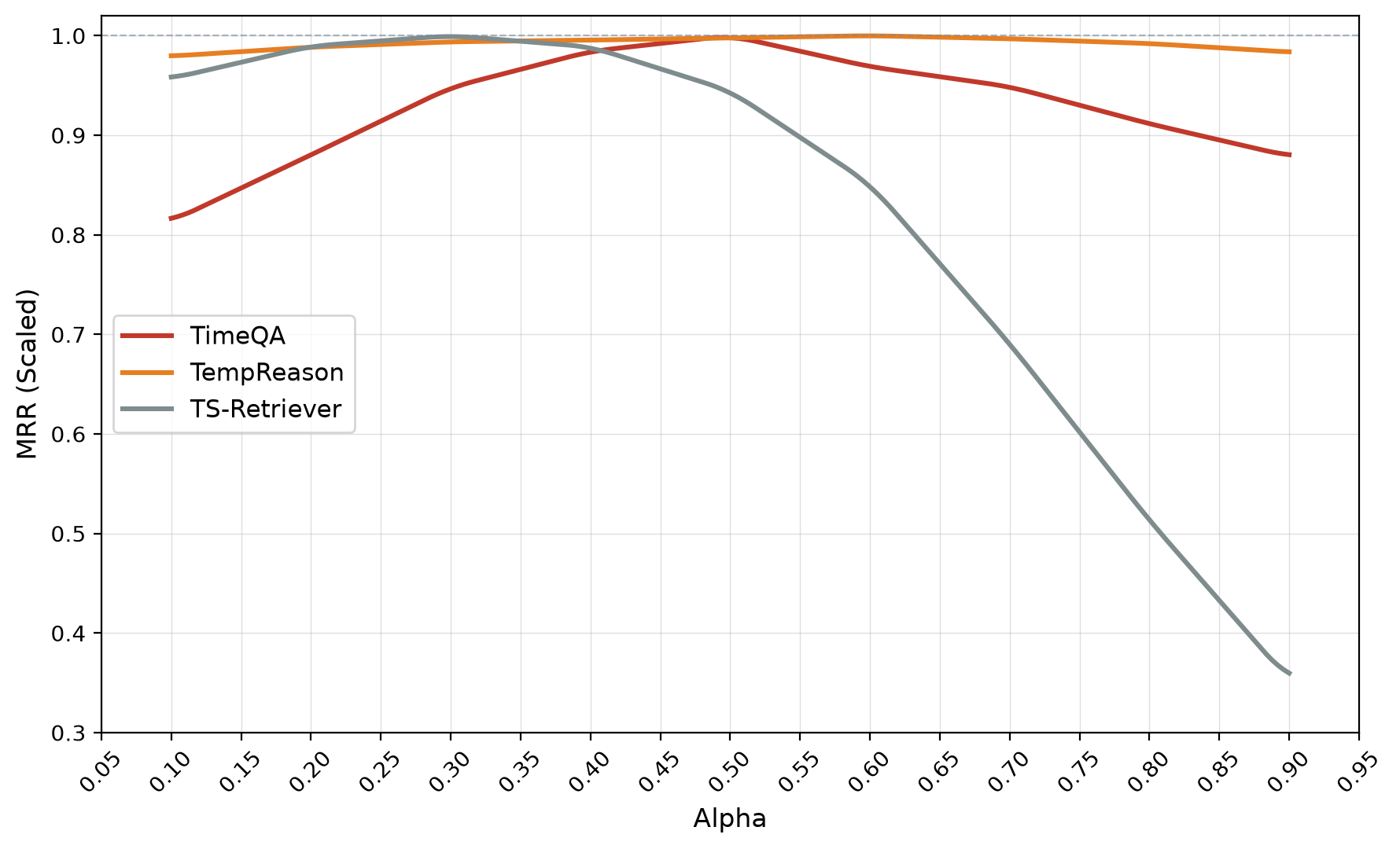}
    \caption{Effect of varying the temporal-semantic interpolation weight $\alpha$, with MRR scaled per benchmark to its own maximum. Low values rely mostly on semantic similarity, high values increasingly on the temporal branch. TimeQA and TempReason favor larger temporal weights, whereas TS-Retriever favors a smaller one because semantic discrimination among topically similar candidates remains important.}
    \label{fig:alpha-values}
\end{figure*}

The interpolation weight $\alpha$ controls how much the final retrieval score
relies on the temporal branch relative to the semantic branch. We sweep candidate
values on the validation split of each benchmark and select one value per
benchmark, applied unchanged to all +TEMPS backbones. Figure~\ref{fig:alpha-values}
shows the sensitivity curves over $\alpha \in [0.1, 0.9]$; the endpoints
themselves are reported in Table~\ref{tab:model_performance}, where $\alpha = 0$
is each semantic-only baseline row and $\alpha = 1$ is \textit{TEMPS w/o
Semantic Model}.

The sweep shows that the best fusion weight is not universal across temporal IR settings, and that the three benchmarks differ as much in the \emph{shape} of the curve as in the location of the optimum. TS-Retriever peaks near $\alpha=0.3$ and falls away sharply above it, losing more than half its MRR by $\alpha=0.85$: its candidates are semantically similar by construction, so the temporal branch is a useful corrective signal but must not dominate the semantic score. TimeQA rises steeply from $\alpha=0.1$, peaks around the middle of the range, and declines gently thereafter. TempReason is nearly flat, staying within $2\%$ of its maximum across the entire sweep, which is consistent with the small +TEMPS margins on that benchmark in Table~\ref{tab:model_performance}.

\subsection{Explicit-Timestamp Oracle Study}
\label{sec:appendix-oracle}

To test whether the TempReason gains reduce to matching an explicit year, we
built two oracles on the held-out test split, restricted to the $4{,}306$ queries
that carry an explicit year: an exact-year regex and a query-interval regex, each
filtering candidates before a semantic rerank with E5\textsubscript{base}-v2.

\begin{table}[t]
\centering
\small
\begin{tabular}{lcc}
\toprule
\textbf{Retriever} & \textbf{MRR} & \textbf{Gold@1} \\
\midrule
Semantic E5\textsubscript{base}-v2        & 0.8332 & 0.703 \\
Exact-year oracle + semantic rerank       & 0.8425 & 0.716 \\
Query-interval oracle + semantic rerank   & 0.8425 & 0.716 \\
Temporal-only (TEMPS head, $\alpha=1$)    & 0.8695 & 0.758 \\
\textbf{TEMPS hybrid ($\alpha=0.8$)}      & \textbf{0.8828} & \textbf{0.780} \\
\bottomrule
\end{tabular}
\caption{Explicit-timestamp oracle study on the held-out TempReason split,
restricted to the $4{,}306$ queries that carry an explicit year. Both oracles
filter candidates by a year regex before a semantic rerank. The temporal head
alone already outperforms both, so the gain does not reduce to year matching.}
\label{tab:oracle}
\end{table}

Both oracles help, and only slightly, lifting MRR by $0.009$ and Gold@1 by
$0.013$ over the semantic baseline. The temporal head alone reaches $0.8695$ MRR
before any semantic fusion, above both oracles, and the hybrid reaches $0.8828$.

Coverage explains the gap. Every query in this subset carries a year, yet in
$2{,}072$ of them ($48\%$) the gold passage contains no year drawn from the
query's interval, because the answer sits at a coarser granularity, is an open
interval, or is paraphrased. Of the $724$ queries TEMPS lifts to rank 1, the
interval oracle still fails on $668$ ($92.3\%$), and $481$ ($66\%$) have a gold
passage with no query-interval year at all. Those are precisely the cases a regex
cannot reach.

\subsection{Computational Resources}

All experiments were conducted on an HPC partition. Each compute node provides two AMD EPYC 7543 CPUs (64 physical cores, 128 threads in total), 512\,GiB of DDR4 memory, and eight NVIDIA A100-SXM4 GPUs with 80\,GB of HBM2 memory each, interconnected via NVLink and NVSwitch. Training was performed on a single node with exclusive allocation, distributing the workload across all eight A100 GPUs through PyTorch's distributed launcher (\texttt{torchrun}, one process per GPU) for a batch size of 2{,}048. Training the temporal module for three epochs over the full corpus took approximately 4 wall-clock hours on the eight A100 GPUs. Training and evaluation scripts, the trained temporal module, and the training-data generation pipeline are released at \url{https://github.com/aldebaran-care/TEMPS}.

\subsection{Artifacts, Licenses, and Intended Use}
\label{sec:appendix-artifacts}

All datasets (TimeQA~\cite{chen2021dataset}, TempReason~\cite{tan2023towards}, TS-Retriever~\cite{wu2024time}, and the Multilingual MLM Temporal Tagging Resources~\cite{lange2022multilingual}) and pretrained models (all-MiniLM-L6-v2, MPNet\textsubscript{base}-v2~\cite{mpnet}, E5\textsubscript{base}-v2~\cite{e5}, BGE\textsubscript{large}-v1.5~\cite{bge}, SFR-Embedding-Mistral, and Qwen2.5-7B-Instruct~\cite{qwen2024qwen25}) used in this work are publicly released for research, and we use them strictly within that intended research scope. Our synthetic temporal data is generated from templates and the publicly available Wikipedia-derived Multilingual MLM Temporal Tagging Resources~\cite{lange2022multilingual}. %
We release our code and the TEMPS temporal module under the MIT license, intended solely for research on temporal retrieval.

\subsection{Statistical Significance}
\label{appendix:significance}

For every (benchmark, baseline) cell of the main results table we compare the
semantic baseline alone ($\alpha=0$) against the same baseline fused with the
trained TEMPS temporal model at the validation-selected $\alpha$ for that
benchmark (TimeQA $\alpha\!=\!0.6$, TempReason $\alpha\!=\!0.8$, TS-Retriever
$\alpha\!=\!0.3$). These values come from the validation split only and are not
varied by semantic backbone or by metric. Per-query metric scores are computed
from the cached retrieval similarities on the test split; the hybrid follows the
per-query min-max-normalized merge
$\alpha\cdot s_{\mathrm{temporal}} + (1-\alpha)\cdot s_{\mathrm{semantic}}$.
We report the two-sided paired Wilcoxon signed-rank test as the primary test, the
two-sided paired $t$-test as a secondary test, and a 95\% bootstrap confidence
interval on the mean per-query gain ($B = 10{,}000$ resamples, seed 42).

\paragraph{What the 48 comparisons show.}
The picture is not uniform across benchmarks, and reporting it as though it were
would misrepresent the tables.

On \textbf{TimeQA}, all 16 comparisons (4 baselines $\times$ 4 metrics) are
positive and significant, with a largest $p$ of $1.787\!\times\!10^{-3}$
(Mistral, Recall@5 and Precision@5). On \textbf{TS-Retriever}, all 16 are
likewise positive and significant, largest $p = 4.412\!\times\!10^{-3}$
(E5\textsubscript{base}-v2, MRR).

\textbf{TempReason} is mixed. Four comparisons are significantly positive
(MPNet\textsubscript{base}-v2 on MRR and NDCG@5, BGE\textsubscript{large}-v1.5 on
MRR, Mistral on MRR), three are not significant at the $0.05$ level
(MPNet\textsubscript{base}-v2 Precision@5, $p = 0.4643$; E5\textsubscript{base}-v2
MRR, $p = 0.4349$; BGE\textsubscript{large}-v1.5 NDCG@5, $p = 0.9555$), and nine
are significantly \emph{negative}, concentrated in Recall@5 and Precision@5. At
$\alpha = 0.8$ the fused score is dominated by the temporal branch, which
reorders the head of the ranking in our favor while costing coverage in the
top-5 set. Section~\ref{sec:experiments} discusses why this benchmark behaves
differently from the other two.

Across all 48 comparisons: 36 are significantly positive, 9 significantly
negative, and 3 are not significant. Applying a Bonferroni correction over all 48
($\alpha = 0.05/48 \approx 1.04\!\times\!10^{-3}$), 32 of the 36 positive results
survive; the four that do not are TimeQA Mistral Recall@5 and Precision@5,
TS-Retriever E5\textsubscript{base}-v2 MRR, and TempReason Mistral MRR.

\paragraph{Per-metric results.}
Full per-metric statistics for TimeQA, TempReason, and TS-Retriever are in
Tables~\ref{tab:sig-timeqa}-\ref{tab:sig-tsretriever}. All metrics are computed
on the held-out test split.

\begin{table*}[t]
\centering
\footnotesize
\caption{TimeQA per-cell statistics ($\alpha = 0.6$, $n = 710$ test queries).
Two-sided paired Wilcoxon signed-rank test, paired $t$-test, and 95\% bootstrap
CI on the mean per-query gain.}
\label{tab:sig-timeqa}
\begin{tabular}{llccccrr}
\toprule
Baseline & Metric & Base & +TEMPS & $\Delta$ & 95\% CI & Wilcoxon $p$ & $t$-test $p$ \\
\midrule
MPNet$_{\text{base}}$-v2 & MRR          & 0.3008 & 0.5043 & $+0.2035$ & $[0.1735, 0.2346]$ & $1.149\!\times\!10^{-32}$ & $2.303\!\times\!10^{-34}$ \\
MPNet$_{\text{base}}$-v2 & NDCG@5       & 0.3121 & 0.5320 & $+0.2199$ & $[0.1873, 0.2532]$ & $2.035\!\times\!10^{-32}$ & $4.937\!\times\!10^{-35}$ \\
MPNet$_{\text{base}}$-v2 & Recall@5     & 0.4803 & 0.6915 & $+0.2113$ & $[0.1704, 0.2535]$ & $3.417\!\times\!10^{-21}$ & $1.704\!\times\!10^{-22}$ \\
MPNet$_{\text{base}}$-v2 & Precision@5  & 0.0961 & 0.1383 & $+0.0423$ & $[0.0338, 0.0507]$ & $3.417\!\times\!10^{-21}$ & $1.704\!\times\!10^{-22}$ \\
\midrule
E5$_{\text{base}}$-v2    & MRR          & 0.3334 & 0.5339 & $+0.2006$ & $[0.1729, 0.2274]$ & $1.288\!\times\!10^{-36}$ & $9.867\!\times\!10^{-41}$ \\
E5$_{\text{base}}$-v2    & NDCG@5       & 0.3712 & 0.5648 & $+0.1936$ & $[0.1653, 0.2215]$ & $2.960\!\times\!10^{-31}$ & $1.091\!\times\!10^{-36}$ \\
E5$_{\text{base}}$-v2    & Recall@5     & 0.6028 & 0.7225 & $+0.1197$ & $[0.0831, 0.1549]$ & $3.313\!\times\!10^{-10}$ & $1.931\!\times\!10^{-10}$ \\
E5$_{\text{base}}$-v2    & Precision@5  & 0.1206 & 0.1445 & $+0.0239$ & $[0.0166, 0.0310]$ & $3.313\!\times\!10^{-10}$ & $1.931\!\times\!10^{-10}$ \\
\midrule
BGE$_{\text{large}}$-v1.5 & MRR         & 0.2955 & 0.5298 & $+0.2343$ & $[0.2059, 0.2623]$ & $1.349\!\times\!10^{-43}$ & $3.290\!\times\!10^{-50}$ \\
BGE$_{\text{large}}$-v1.5 & NDCG@5      & 0.3269 & 0.5566 & $+0.2296$ & $[0.1996, 0.2602]$ & $1.226\!\times\!10^{-35}$ & $1.201\!\times\!10^{-42}$ \\
BGE$_{\text{large}}$-v1.5 & Recall@5    & 0.5465 & 0.7070 & $+0.1606$ & $[0.1197, 0.2014]$ & $9.165\!\times\!10^{-14}$ & $3.039\!\times\!10^{-14}$ \\
BGE$_{\text{large}}$-v1.5 & Precision@5 & 0.1093 & 0.1414 & $+0.0321$ & $[0.0239, 0.0403]$ & $9.165\!\times\!10^{-14}$ & $3.039\!\times\!10^{-14}$ \\
\midrule
Mistral                   & MRR          & 0.4063 & 0.5360 & $+0.1297$ & $[0.1006, 0.1596]$ & $4.295\!\times\!10^{-15}$ & $1.627\!\times\!10^{-16}$ \\
Mistral                   & NDCG@5       & 0.4455 & 0.5622 & $+0.1168$ & $[0.0867, 0.1476]$ & $1.120\!\times\!10^{-12}$ & $2.026\!\times\!10^{-13}$ \\
Mistral                   & Recall@5     & 0.6549 & 0.7113 & $+0.0563$ & $[0.0211, 0.0915]$ & $1.787\!\times\!10^{-3}$ & $1.742\!\times\!10^{-3}$ \\
Mistral                   & Precision@5  & 0.1310 & 0.1423 & $+0.0113$ & $[0.0042, 0.0186]$ & $1.787\!\times\!10^{-3}$ & $1.742\!\times\!10^{-3}$ \\
\bottomrule
\end{tabular}
\end{table*}

\begin{table*}[t]
\centering
\footnotesize
\caption{TempReason per-cell statistics ($\alpha = 0.8$, $n = 7{,}518$ test
queries). Two-sided paired Wilcoxon signed-rank test, paired $t$-test, and 95\%
bootstrap CI on the mean per-query gain.}
\label{tab:sig-tempreason}
\begin{tabular}{llccccrr}
\toprule
Baseline & Metric & Base & +TEMPS & $\Delta$ & 95\% CI & Wilcoxon $p$ & $t$-test $p$ \\
\midrule
MPNet$_{\text{base}}$-v2 & MRR          & 0.4957 & 0.6202 & $+0.1246$ & $[0.1168, 0.1323]$ & $2.924\!\times\!10^{-187}$ & $3.742\!\times\!10^{-207}$ \\
MPNet$_{\text{base}}$-v2 & NDCG@5       & 0.5420 & 0.6283 & $+0.0863$ & $[0.0786, 0.0940]$ & $3.026\!\times\!10^{-85}$ & $1.267\!\times\!10^{-105}$ \\
MPNet$_{\text{base}}$-v2 & Recall@5     & 0.7936 & 0.7732 & $-0.0204$ & $[-0.0301, -0.0109]$ & $7.821\!\times\!10^{-10}$ & $3.005\!\times\!10^{-5}$ \\
MPNet$_{\text{base}}$-v2 & Precision@5  & 0.1724 & 0.1724 & $-0.0001$ & $[-0.0022, 0.0020]$ & $0.4643$ & $0.9415$ \\
\midrule
E5$_{\text{base}}$-v2    & MRR          & 0.6208 & 0.6278 & $+0.0070$ & $[0.0010, 0.0130]$ & $0.4349$ & $0.02344$ \\
E5$_{\text{base}}$-v2    & NDCG@5       & 0.6604 & 0.6364 & $-0.0239$ & $[-0.0300, -0.0178]$ & $3.482\!\times\!10^{-19}$ & $1.494\!\times\!10^{-14}$ \\
E5$_{\text{base}}$-v2    & Recall@5     & 0.8587 & 0.7798 & $-0.0790$ & $[-0.0875, -0.0707]$ & $1.896\!\times\!10^{-76}$ & $9.434\!\times\!10^{-75}$ \\
E5$_{\text{base}}$-v2    & Precision@5  & 0.1893 & 0.1738 & $-0.0154$ & $[-0.0171, -0.0137]$ & $2.067\!\times\!10^{-68}$ & $5.844\!\times\!10^{-68}$ \\
\midrule
BGE$_{\text{large}}$-v1.5 & MRR         & 0.5937 & 0.6286 & $+0.0349$ & $[0.0284, 0.0413]$ & $3.228\!\times\!10^{-20}$ & $5.702\!\times\!10^{-26}$ \\
BGE$_{\text{large}}$-v1.5 & NDCG@5      & 0.6327 & 0.6364 & $+0.0037$ & $[-0.0027, 0.0102]$ & $0.9555$ & $0.2514$ \\
BGE$_{\text{large}}$-v1.5 & Recall@5    & 0.8413 & 0.7781 & $-0.0632$ & $[-0.0715, -0.0548]$ & $2.733\!\times\!10^{-55}$ & $3.768\!\times\!10^{-49}$ \\
BGE$_{\text{large}}$-v1.5 & Precision@5 & 0.1851 & 0.1735 & $-0.0116$ & $[-0.0134, -0.0099]$ & $4.774\!\times\!10^{-40}$ & $2.350\!\times\!10^{-38}$ \\
\midrule
Mistral                   & MRR          & 0.6191 & 0.6320 & $+0.0128$ & $[0.0050, 0.0206]$ & $6.075\!\times\!10^{-3}$ & $1.250\!\times\!10^{-3}$ \\
Mistral                   & NDCG@5       & 0.6731 & 0.6429 & $-0.0302$ & $[-0.0379, -0.0223]$ & $1.097\!\times\!10^{-17}$ & $1.698\!\times\!10^{-14}$ \\
Mistral                   & Recall@5     & 0.8967 & 0.7887 & $-0.1081$ & $[-0.1173, -0.0988]$ & $4.780\!\times\!10^{-118}$ & $2.467\!\times\!10^{-112}$ \\
Mistral                   & Precision@5  & 0.1959 & 0.1756 & $-0.0203$ & $[-0.0222, -0.0183]$ & $1.522\!\times\!10^{-94}$ & $1.505\!\times\!10^{-92}$ \\
\bottomrule
\end{tabular}
\end{table*}

\begin{table*}[t]
\centering
\footnotesize
\caption{TS-Retriever per-cell statistics ($\alpha = 0.3$, $n = 2{,}595$ test
queries). Two-sided paired Wilcoxon signed-rank test, paired $t$-test, and 95\%
bootstrap CI on the mean per-query gain.}
\label{tab:sig-tsretriever}
\begin{tabular}{llccccrr}
\toprule
Baseline & Metric & Base & +TEMPS & $\Delta$ & 95\% CI & Wilcoxon $p$ & $t$-test $p$ \\
\midrule
MPNet$_{\text{base}}$-v2 & MRR          & 0.3902 & 0.5596 & $+0.1693$ & $[0.1576, 0.1811]$ & $1.080\!\times\!10^{-173}$ & $6.315\!\times\!10^{-155}$ \\
MPNet$_{\text{base}}$-v2 & NDCG@5       & 0.2396 & 0.3892 & $+0.1496$ & $[0.1415, 0.1581]$ & $3.746\!\times\!10^{-195}$ & $1.116\!\times\!10^{-217}$ \\
MPNet$_{\text{base}}$-v2 & Recall@5     & 0.2377 & 0.3760 & $+0.1383$ & $[0.1284, 0.1483]$ & $3.059\!\times\!10^{-140}$ & $3.438\!\times\!10^{-144}$ \\
MPNet$_{\text{base}}$-v2 & Precision@5  & 0.1645 & 0.2687 & $+0.1042$ & $[0.0980, 0.1104]$ & $2.268\!\times\!10^{-157}$ & $2.212\!\times\!10^{-197}$ \\
\midrule
E5$_{\text{base}}$-v2    & MRR          & 0.7880 & 0.8025 & $+0.0145$ & $[0.0036, 0.0252]$ & $4.412\!\times\!10^{-3}$ & $7.055\!\times\!10^{-3}$ \\
E5$_{\text{base}}$-v2    & NDCG@5       & 0.6431 & 0.6726 & $+0.0295$ & $[0.0219, 0.0372]$ & $2.147\!\times\!10^{-13}$ & $8.938\!\times\!10^{-14}$ \\
E5$_{\text{base}}$-v2    & Recall@5     & 0.6122 & 0.6427 & $+0.0305$ & $[0.0226, 0.0381]$ & $1.936\!\times\!10^{-14}$ & $2.765\!\times\!10^{-14}$ \\
E5$_{\text{base}}$-v2    & Precision@5  & 0.4238 & 0.4477 & $+0.0239$ & $[0.0183, 0.0294]$ & $3.706\!\times\!10^{-15}$ & $3.447\!\times\!10^{-17}$ \\
\midrule
BGE$_{\text{large}}$-v1.5 & MRR         & 0.7151 & 0.7936 & $+0.0785$ & $[0.0677, 0.0893]$ & $1.714\!\times\!10^{-44}$ & $7.336\!\times\!10^{-44}$ \\
BGE$_{\text{large}}$-v1.5 & NDCG@5      & 0.5459 & 0.6463 & $+0.1004$ & $[0.0919, 0.1090]$ & $4.944\!\times\!10^{-106}$ & $4.793\!\times\!10^{-108}$ \\
BGE$_{\text{large}}$-v1.5 & Recall@5    & 0.5163 & 0.6100 & $+0.0938$ & $[0.0841, 0.1035]$ & $7.248\!\times\!10^{-77}$ & $1.776\!\times\!10^{-78}$ \\
BGE$_{\text{large}}$-v1.5 & Precision@5 & 0.3600 & 0.4250 & $+0.0650$ & $[0.0587, 0.0713]$ & $1.368\!\times\!10^{-67}$ & $2.360\!\times\!10^{-88}$ \\
\midrule
Mistral                   & MRR          & 0.7539 & 0.8240 & $+0.0700$ & $[0.0604, 0.0798]$ & $1.460\!\times\!10^{-41}$ & $1.168\!\times\!10^{-42}$ \\
Mistral                   & NDCG@5       & 0.6111 & 0.6892 & $+0.0781$ & $[0.0704, 0.0856]$ & $9.911\!\times\!10^{-82}$ & $4.425\!\times\!10^{-84}$ \\
Mistral                   & Recall@5     & 0.5856 & 0.6471 & $+0.0615$ & $[0.0532, 0.0697]$ & $2.155\!\times\!10^{-47}$ & $1.311\!\times\!10^{-47}$ \\
Mistral                   & Precision@5  & 0.4058 & 0.4508 & $+0.0450$ & $[0.0395, 0.0506]$ & $4.573\!\times\!10^{-41}$ & $1.538\!\times\!10^{-54}$ \\
\bottomrule
\end{tabular}
\end{table*}

\clearpage
\raggedbottom
\section{Qualitative Analysis: What Moves and Why}
\label{sec:appendix-qualitative}

TimeQA is the right benchmark for this analysis because the candidates for a
given query are drawn from a single Wikipedia article. They are close to
identical in topic and differ mainly in temporal scope, so a change in ranking
cannot easily be attributed to better topical matching. For every query on the
held-out test split ($n=710$, one gold paragraph each) we compare three
rankings: the semantic-only baseline (E5\textsubscript{base}-v2, $\alpha=0$), the
temporal-only variant ($\alpha=1$), and the fused +TEMPS ranking at the selected
$\alpha=0.6$.

\paragraph{How often does the gold passage move, and in which direction?}
Table~\ref{tab:qualitative-movement} counts a \textit{promotion} when the gold
passage enters rank 1 (or the top 5) under +TEMPS having been outside it under
the semantic baseline, and a \textit{demotion} when the reverse happens.

\begin{table}[t]
\centering
\small
\begin{tabular}{lccc}
\toprule
\textbf{Movement} & \textbf{Promoted} & \textbf{Demoted} & \textbf{Ratio} \\
\midrule
Into rank 1 & 206 & 31 & $6.6{:}1$ \\
Into top 5  & 134 & 49 & $2.7{:}1$ \\
\bottomrule
\end{tabular}
\caption{Movement of the gold passage on the held-out TimeQA test split
($n=710$) when TEMPS is added to E5\textsubscript{base}-v2 at $\alpha=0.6$.}
\label{tab:qualitative-movement}
\end{table}

Adding TEMPS moves the gold passage to rank 1 more than six times as often as it
displaces one, and into the top 5 close to three times as often as it drops one.
In aggregate, top-1 accuracy rises from $0.134$ ($95/710$) to $0.380$
($270/710$), a factor of $2.8$, while Gold@5 rises from $0.603$ to $0.723$; the
latter two figures are the Recall@5 entries for E5\textsubscript{base}-v2 in
Table~\ref{tab:model_performance}. The gains land on same-topic, different-time
cases, which is the class the temporal branch is meant to fix. The
$31$ rank-1 and $49$ top-5 regressions are the cost of that trade.

\paragraph{Three cases.}
Table~\ref{tab:qualitative-cases} gives three queries from TimeQA, quoted
verbatim, where a topically strong distractor buries the gold passage. In each
the baseline's leading hits are the article title and its undated lead
paragraphs, which are topically maximal and carry no temporal information at all,
while the gold passage is a short dated entry far down the ranking. In all three
the temporal-only variant also places the gold passage at rank 1, with no
query-document semantic term in the score.

\begin{table*}[!t]
\centering
\small
\begin{tabular}{p{4.4cm}p{4.4cm}p{4.8cm}}
\toprule
\textbf{Query (candidate pool)} & \textbf{Top semantic hit} & \textbf{Gold passage} \\
\midrule
``Who was the leader of the advisory body Internet Architecture Board from 1993 to 1995?'' (127)
& ``Internet Architecture Board'' -- the article title, no year.
& ``Christian Huitema -- March 1993 to July 1995'': rank 18 $\rightarrow$ \textbf{1}. \\
\midrule
``Coach Joachim Low worked with what team from 1998 to 1999?'' (84)
& His career biography, dominated by the 2014 World Cup and later national-team years.
& His 1 July 1998 move to Fenerbah\c{c}e: rank 11 $\rightarrow$ \textbf{1}. \\
\midrule
``Which queen did the The Arnolfini Marriage belong to from 1530 to 1558?'' (72)
& The painting's creation as a 1434 oil on oak panel.
& The 1530 inheritance by Mary of Hungary, described in the 1558 inventory: rank 17 $\rightarrow$ \textbf{1}. \\
\bottomrule
\end{tabular}
\caption{TimeQA cases where the top semantic hit is topically maximal and
temporally wrong. Candidate-pool size in parentheses; ranks are semantic-only
$\rightarrow$ +TEMPS. In all three the temporal-only variant also places the gold
passage at rank 1.}
\label{tab:qualitative-cases}
\end{table*}

The pattern is the same in each: the baseline's first hit is the most topical
passage in the article and carries no date, while the passage carrying the
requested interval sits well down the ranking. That the temporal-only variant
recovers it without any semantic signal points at the temporal component as the
source of the fix rather than at incidental lexical overlap. These three were
selected as illustrative wins, so they show the mechanism rather than the average
case; Table~\ref{tab:qualitative-movement} gives the aggregate picture, including
the regressions.

\end{document}